\documentclass[a4paper, 11pt]{article}
\usepackage{geometry}
\usepackage{titlesec}
\titleformat{\section}{\normalfont\large\bfseries}{\thesection}{1em}{}
\titleformat{\subsection}{\normalfont\normalsize\bfseries}{\thesubsection}{1em}{}

\titlespacing\section{0pt}{12pt plus 3pt minus 3pt}{3pt plus 1pt minus 1pt}
\titlespacing\subsection{0pt}{10pt plus 3pt minus 3pt}{3pt plus 1pt minus 1pt}
\titlespacing\subsubsection{0pt}{8pt plus 3pt minus 3pt}{3pt plus 1pt minus 1pt}

\usepackage[numbers,round,sort&compress]{natbib}
\usepackage{microtype}
\usepackage{mathtools,amsfonts,dsfont,amsmath,amssymb}
\usepackage{tabularx,multirow,array}
\usepackage{caption, subcaption}
\usepackage{xcolor}		
\definecolor{mycolor}{RGB}{74,168,142}
\usepackage[colorlinks=true, linkcolor=mycolor, urlcolor=mycolor, citecolor=mycolor, anchorcolor=mycolor]{hyperref}
\usepackage{rotating}
\usepackage[algo2e]{algorithm2e}
\usepackage{algorithm, algpseudocode}

\newcommand*{\tens}[1]{\boldsymbol{#1}}

\usepackage[final]{changes}
\begin{document} 

\title{\Large \bf Intrinsic fluctuations of reinforcement learning promote cooperation}
\author{Wolfram Barfuss$^{*1}$ and Janusz M. Meylahn$^{*2,3}$\\
\footnotesize{wolfram.barfuss@gmail.com $|$ j.m.meylahn@utwente.nl (corresponding author)}}

\date{$^1$University of Tübingen,\added{ Tübingen AI Center,} GER\\
    $^2$Department of Applied Mathematics, University of Twente, NL\\
    $^3$Dutch Institute of Emergent Phenomena, University of Amsterdam, NL\\
    \footnotesize{$^*$Both authors contributed equally to this work} }
\maketitle
\thispagestyle{empty}


{\flushleft{\bf Abstract.}} 
In this work, we ask for and answer what makes classical \added{temporal-difference} reinforcement learning \added{with $\epsilon$-greedy strategies} cooperative.
Cooperating in social dilemma situations is vital for animals, humans, and machines. While evolutionary theory revealed a range of mechanisms promoting cooperation, the conditions under which agents learn to cooperate are contested. Here, we demonstrate which and how individual elements of the multi-agent learning setting lead to cooperation. \deleted{Specifically, we consider the widely used temporal-difference reinforcement learning algorithm with epsilon-greedy exploration in the classic environment of an}\added{We use the} iterated Prisoner's dilemma with one-period memory\added{ as a testbed}. Each of the two learning agents learns a strategy that conditions the following action choices on both agents' action choices of the last round. 
We find that next to a high caring for future rewards, a low exploration rate, and a small learning rate, it is primarily intrinsic stochastic fluctuations of the reinforcement learning process which double the final rate of cooperation to up to 80\%. Thus, inherent noise is not a necessary evil of the iterative learning process. It is a critical asset for the learning of cooperation. 
However, we also point out the trade-off between a high likelihood of cooperative behavior and achieving this in a reasonable amount of time. Our findings are relevant for purposefully designing cooperative algorithms and regulating undesired collusive effects.



\section*{Introduction}

Problems of cooperation are ubiquitous and essential, for biological phenomena, as in the evolution of cooperation under natural selection, for human behavior, such as in cartel pricing or traffic, and increasingly so for intelligent machines with automated trading and self-driving cars \cite{DafoeEtAl2021, BertinoEtAl2020, Levin2020}.
In social dilemmas, individual incentives and collective welfare are not aligned. Individuals profit from exploiting others or fear being exploited by others, while at the same time, the collective welfare is maximized if all choose to cooperate \cite{Dawes1980}.

\added{Understanding the conditions under which self-learning agents learn to cooperate spontaneously - without explicit intent to do so -   is critical for three reasons:
1) It provides an alternative route to the emergence of (human and animal) cooperation when an evolutionary explanation is unlikely. 
2) It guides the design of intelligent self-learning algorithms, which are supposed to be cooperative.
3) It provides policymakers and regulators the necessary background to design novel anti-trust legislation against undesirable collusion, e.g., in algorithmic pricing situations, where not doing so could lead to significant loss of consumer welfare \cite{Harrington2018}.}

While evolutionary theory revealed a range of mechanisms that promote cooperative behavior \deleted{\cite{AxelrodHamilton1981, NowakSigmund1993, Nowak2006}}\added{, from direct and indirect reciprocity to spatial and network effects \cite{AxelrodHamilton1981, NowakSigmund1993, Nowak2006, PercEtAl2013, PercEtAl2017}}, the conditions under which individually learning agents learn to cooperate are contested.
Some works suggest that independent reinforcement learning agents are capable of spontaneously cooperating without explicit intent to do so \cite{MasudaOhtsuki2009, EzrachiStucke2016, CiminiSanchez2014, EzakiEtAl2016, EzrachiStucke2017, PerolatEtAl2017, LeiboEtAl2017, KuhnTadelis2017, Calvano2019a, BarbosaEtAl2020}.
Other works argue that the emergence of cooperation from independent learning agents is unlikely \cite{SandholmCrites1996, Schrepel2017, Schwalbe2018, PeysakhovichLerer2018, DafoeEtAl2020}, and therefore specific algorithmic features are required to promote cooperation \cite{PeysakhovichLerer2018a, LererPeysakhovich2018, FoersterEtAl2018, HughesEtAl2018, EcclesEtAl2019, Baker2020, WangEtAl2019, HughesEtAl2020, Meylahn2022}\added{\cite{BowlingVeloso2002, deCoteEtAl2006, PanaitEtAl2006, StimpsonGoodrich2003}}.
\added{As such, reinforcement learning variants called \textit{aspiration learning}, which go back to a seminal work in psychology from Bush and Mosteller \cite{BushMosteller1951}, have been extensively investigated in social dilemmas. Whether two co-players learn to cooperate depends on the (dynamics of the) aspiration level \cite{MacyFlache2002, IzquierdoEtAl2008, MasudaNakamura2011}. This finding has been confirmed and extended to spatial or networked social dilemmas \cite{ZhangEtAl2012, JiaMa2013, JiaEtAl2021, SongEtAl2022}. 
Aspiration learning has also been found to explain human play in behavioral experiments well \cite{CiminiSanchez2014, EzakiEtAl2016}. However, comparably little is known about whether, when, and how cooperative behavior spontaneously emerges from the reinforcement learning variants called \textit{temporal-difference learning}, which are extensively used in machine learning applications and is the dominant model used to explain neuroscientific experiments \cite{BotvinickEtAl2020}}.

\deleted{Understanding the conditions under which individually, self-learning agents learn to cooperate spontaneously - without explicit intent to do so -   is critical for three reasons:
1) It provides an alternative route to the emergence of (human and animal) cooperation when an evolutionary explanation is unlikely. 
2) It guides the design of intelligent self-learning algorithms, which are supposed to be cooperative.
3) It provides policymakers and regulators the necessary background to design novel anti-trust legislation against undesirable collusion, e.g., in algorithmic pricing situations, where not doing so could lead to significant loss of consumer welfare \cite{Harrington2018}.}

\begin{figure*}
	\centering
    \begin{subfigure}[b]{0.61\linewidth}
        \centering
	    \includegraphics[width=0.99\linewidth]{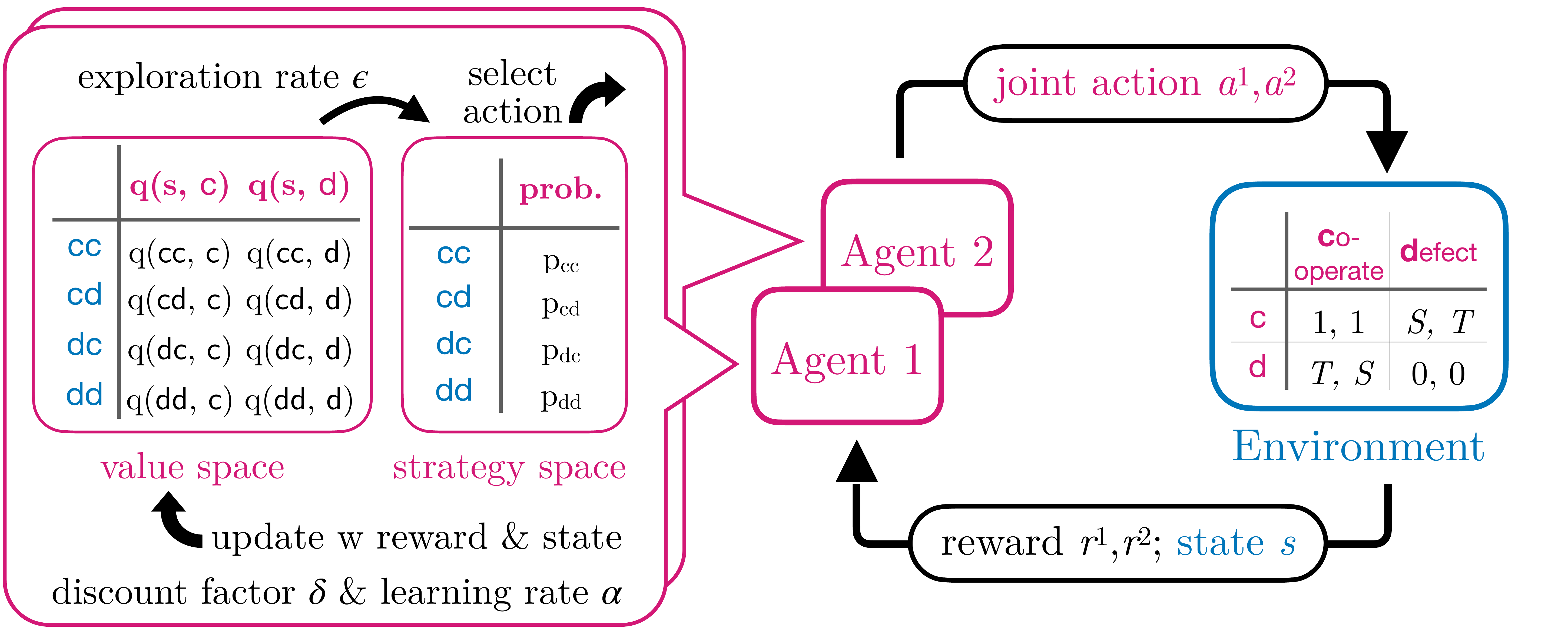}
    	\caption{Agent-Environment Interface}
	\end{subfigure}
    \begin{subfigure}[b]{0.38\linewidth}
        \centering
	    \includegraphics[width=0.99\linewidth]{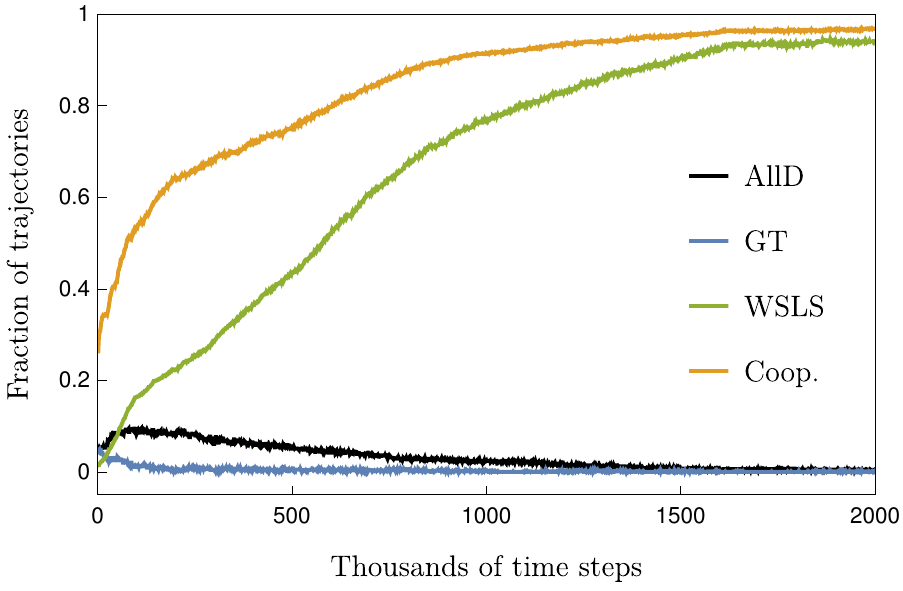}
        \caption{Strategies and cooperation trajectories}
	\end{subfigure}
	\caption{\textbf{Overview.} (a) Model sketch. (b) Fraction of 1000 samples from random initial state-action values in each of the three equilibria (All-Defect (AllD) in black, Grim-Trigger (GT) in blue, and Win-Stay, Lose-Shift (WSLS) in green) as a function of time when using $\epsilon$-greedy temporal-difference learning with $\epsilon=0.01, \delta=0.98$ $\alpha=0.1$. The fraction of times that both Q-learners cooperated in the last thousand time periods averaged over 1000 sample trajectories in light gray. Environment parameters are $T=1.5$ and $S=-0.2$.}
	\label{fig:Model}
\end{figure*}

In this work, we ask when and how cooperative behavior spontaneously emerges from \added{temporal-difference learning with $\epsilon$-greedy strategy functions. This question is motivated by recent work on algorithmic collusion \cite{Calvano2020} and the fact that $\epsilon$-greedy strategies are frequently used in machine learning\cite{sutton2018reinforcement}}\deleted{a foundational reinforcement learning process extensively used in machine learning applications and to explain neuroscientific experiments}.
The problem is that reinforcement learning is typically highly stochastic and data-inefficient, making it challenging to understand which features are decisive for learning cooperation. We solve this problem by dissecting the reinforcement learning processes into three parts using multiple mathematical techniques.

First, we consider the \textit{stability} of strategies under reinforcement learning. We analytically derive when strategies are stable given how much the agents care for future rewards and explore the environment. Only one out of the three possible, stable strategies supports cooperation robustly.

Second, we consider the \textit{learnability} of this equilibrium. We use deterministic strategy-average learning dynamics to compute the size of \added{the} basin\added{s}-of-attraction given the agents' learning rate. We find a maximum of approx. 40-50\% of robust cooperation.

Third and last, we consider the \textit{stochasticity} of the learning process. We simulate a stochastic batch-learning algorithm and find that the  cooperative equilibrium steadily increases to 80-100\%. Thus, a significant fraction of trials reaching the cooperative equilibrium must be due to the inherent fluctuations of the reinforcement learning process.

\subsection*{Learning Algorithm and Environment}

We consider the generalized and advantageous temporal-difference reinforcement learning algorithm \textit{Expected SARSA} \cite{Rummery1994, Sutton1995, sutton2018reinforcement} with $\epsilon$-greedy exploration. 
At each time step $t$, agent $i \in \{1,2\}$ chooses between two possible actions, $a \in \mathcal{A}^1 = \mathcal{A}^2 = \{\textsf{c}, \textsf{d}\}$, which represent a cooperative or a defective act.
Given the joint action $\boldsymbol a = \{a^1, a^2\}$ and the current state of the environment $s \in \mathcal S$, each agent receives a payoff or reward $r^i(s, \tens a)$ and the environment transitions to a new state $s' \in \mathcal S$ with probability $p(s'| \tens a, s)$.
%
Agent $i$ chooses action $a$ with frequency $x^i_t(a|s)$ which depends on the current environmental state $s \in \mathcal S$.

Agents derive these frequencies $x^i_t(a|s)$ from their state-action values $q^i_t(s, a)$ according to the $\epsilon$-greedy exploration scheme. Each agent selects the action with the largest state-action value with probability $1-\epsilon$, and with probability $\epsilon$, it selects an action uniformly at random. For the two-action case,
\begin{equation}
	x^i_t(\mathsf{c}|s) =
		\begin{cases}
			1-\epsilon/2 &\text{if }q^i_t(\mathsf{c},a) > q^i_t(\mathsf{d},a)\\
			\epsilon/2 &\text{otherwise }
		\end{cases}.
\label{eq:EpsGreedyStrategy}
\end{equation}
$x^i_t(\mathsf{d}|s)$ is defined analogously. The parameter $\epsilon$ regulates the exploration-exploitation trade-off.

The state-action values are updated after each time step as,
\begin{align}
	\label{eq:Qupdate} 
	q^i_{t+1}(s_t, a_t) =& (1-\alpha)q^i_{t}(s_t, a_t) +\alpha\Big[r^i_t+\delta \sum_a x^i_t(a|s_{t+1}) q^i_t(s_{t+1}, a) \Big],
\end{align}
where the parameter $\alpha \in [0,1]$ denotes the learning rate, $r^i_t = r^i(s_t, \tens a_t)$ denotes the rewards agent $i$ receives at time step $t$ and $\delta \in [0,1)$ denotes the agents' discount factor, regulating how much they care for future rewards. 
For simplicity, we consider homogeneous and constant parameters $\alpha, \epsilon, \delta$ during the learning process.

The environment we study is the iterated Prisoner's Dilemma. It is perhaps the most iconic and straightforward model system to investigate the preconditions for cooperative behavior, with an established body of research in fields as diverse as political science and evolutionary biology \cite{PressDyson2012}. Because of its simplicity in carving out the tensions between individual incentives and collective welfare, we use it here as a model system to highlight an effect that is, therefore, likely to exist in other larger systems that retain similar tensions between individual incentives and collective welfare.
Specifically, we use reward matrices given by,
\begin{center}
	\vspace{-0.125cm}
	\begin{tabular}{cc|c|c|}
		& \multicolumn{1}{c}{} & \multicolumn{2}{c}{Agent $2$}\\
		& \multicolumn{1}{c}{} & \multicolumn{1}{c}{$\mathsf{c}$}  & \multicolumn{1}{c}{$\mathsf{d}$} \\ 
		\cline{3-4}
		\multirow{2}*{Agent $1$}  & $\mathsf{c}$ & $ 1,1 $ & $S,T$ \\\cline{3-4}
		& $\mathsf{d}$ & $T,S$ & $0,0$ \\ 
		\cline{3-4}
	\end{tabular} \qquad\quad\quad{}
	\vspace{0.125cm}
\end{center}
with $T>1>0>S$. 
The rewards for each combination of actions are written in the cells of the matrix. Each cell's first (second) element denotes the payoff for agent 1 (2). With $T>1$ and $S<0$, each agent prefers defection over cooperation, regardless of what the other agent is doing. The dilemma is that both agents could achieve a higher reward if both cooperate.

However, when the game is repeated for multiple rounds, agents can condition their frequencies of choosing actions on the actions of past rounds, and mutual defection is no longer inevitable. A famous example is the Tit-for-Tat strategy \cite{AxelrodHamilton1981}, in which you cooperate if your co-player cooperated, and you defect if your co-player defected in the last round.

We are interested in how two reinforcement learning agents endogenously learn such memory-1 strategies. Therefore, we embed the stateless Prisoner's Dilemma game into an environment where the current environmental state $s_t=a^1_{t-1}a^2_{t-1}$ signals the actions of the last round. Thus, the state set reads $\mathcal S = \{\mathsf{cc}, \mathsf{cd}, \mathsf{dc}, \mathsf{dd}\}$. Fig.~\ref{fig:Model} \!(a) illustrates our setting.

\begin{figure*}
    \centering
    \begin{subfigure}[b]{0.48\textwidth}
        \centering
	    \includegraphics[scale=0.85]{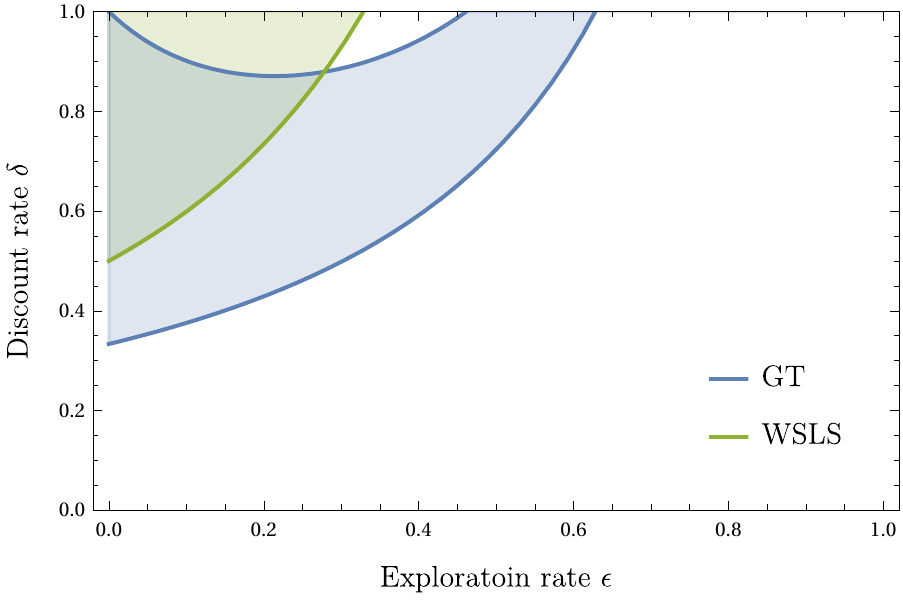}
		\vspace{-0.1cm}
    	\caption{Environment parameters $S=-0.2, T=1.5$ }
	\end{subfigure}
    \begin{subfigure}[b]{0.48\textwidth}
        \centering
    	\includegraphics[scale=0.85]{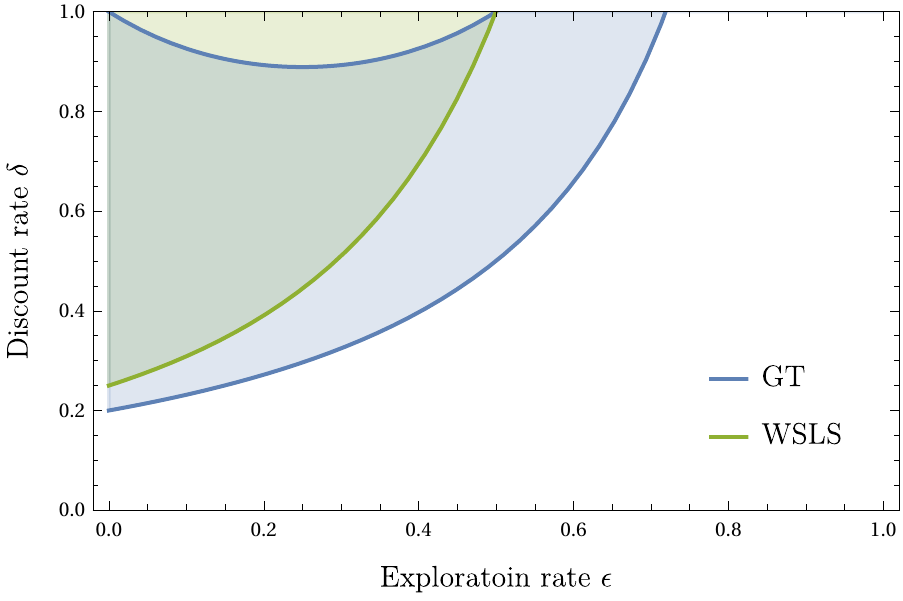}
		\vspace{-0.1cm}
        \caption{Environment parameters $S=-0.25, T=1.25$ }
	\end{subfigure}
	\caption{\textbf{Stability parameter space.} Phase diagrams show which strategy equilibrium solutions are possible. The All-Defect (AllD) solution is possible everywhere. The Grim-Trigger (GT) solution is possible in the blue region, and the Win-Stay, Lose-Shift (WSLS) strategy is possible in the green region.}
	\label{fig:phasediagramdeltaepsilon}
\end{figure*}

Recent work has shown that only three strategy pairs are an equilibrium for $\epsilon$-greedy temporal-difference learning with one-period memory in the iterated Prisoner's Dilemma in the small exploration rate limit \cite{UsuiUeda2021,Meylahn2021}. Interestingly, the Tit-for-Tat strategy is not an equilibrium. The three strategies are All-Defect (AllD), Grim-Trigger (GT), and  Win-Stay, Lose-Shift (WSLS).
In AllD, both agents defect regardless of the previous period.
In GT, agents play AllD except when both players cooperated in the last period, which is answered by cooperation.
And in WSLS, agents play GT, except that cooperation also follows a previous period of both players defecting. Only the WSLS equilibrium leads to robust cooperation. Under the GT strategy, both agents keep cooperating, given that they have cooperated in the last round. But erroneous or exploration moves make it more likely to switch from full cooperation to full defection than switching from full defection back to full cooperation.

Fig.~\ref{fig:Model} \!(b) shows the running average of the fraction of times both players cooperated (yellow) as a function of time, beginning from 1000 random initial state-action values. The  trajectories of the fraction of the three stable strategies are likewise shown. In the end, the agents cooperate  almost always. This shows that temporal-difference reinforcement learning can spontaneously learn to cooperate. However, it leaves open the question of why the agents learn to cooperate and which features of the learning algorithm are decisive for its ability to cooperate. In particular, the effects of the exploration and learning rates, $\epsilon$ and $\alpha$, and the intrinsic stochasticity of the reinforcement learning process remain unclear.

\section*{Dissecting Reinforcement Learning}

To shed light on the questions raised above, we will dissect the reinforcement learning processes into three parts. 
First, we consider the \textit{stability} of strategy pairs under reinforcement learning, considering agents' discount factor $\delta$ and the exploration rate $\epsilon$.
Second, we analyze the \textit{learnability} of this equilibrium, taking into account the learning rate $\alpha$. Third and last, we consider the \textit{stochasticity} of the learning process by introducing a batch-learning variant of our temporal-difference reinforcement algorithm with a batch size parameter $K$.

\paragraph{Stability.}
This section shows how the exploration rate $\epsilon$ affects the stability landscape. We analytically derive when strategy pairs are stable under the reinforcement learning update outside the small exploration limit. To do so, we refine the mathematical technique of \textit{Mutual Best-Response Networks} \cite{Meylahn2021}. With this method, we construct a directed network where the nodes represent the strategy pairs, and the edges represent a best-response relationship (see \textit{Methods}). 

We find that AllD is always a solution. The condition for having WSLS as a solution is 
\begin{equation}
\label{eq:WSLScondition}
    \delta >  \frac{2(T-1) + \epsilon(1-S-T)}{2(1 - \epsilon)^{2}},
\end{equation}
while the condition for having GT as a solution is
\begin{equation}
\label{eq:GTcondition}
    \frac{2 S + \epsilon(1-S-T)}{(1 - \epsilon) [(2-\epsilon)S -\epsilon T]} > \delta > \frac{2(T-1) + \epsilon (1-S-T)}{(1-\epsilon)(2T -\epsilon[S+T])}.
\end{equation}
The condition for WSLS (Eq.~\ref{eq:WSLScondition}) is always greater than the lower bound condition for GT (Eq.~\ref{eq:GTcondition}). This means the robustly cooperative WSLS always requires a higher discount factor than the GT strategy equilibrium.

Fig.~\ref{fig:phasediagramdeltaepsilon} illustrates when the three equilibrium strategy pairs, AllD, GT, and WSLS, are stable, given the discount factor $\delta$ and the exploration rate $\epsilon$. The cooperative WSLS strategy pair is stable when $\delta$ is high, and $\epsilon$ is small. The GT strategy pair also requires a high $\delta$ and a small $\epsilon$ to become stable, yet, with less extreme parameter values. Interestingly, for large discount factors $\delta$, our theory predicts the GT equilibrium to lose stability for exploration rates $\epsilon$ between 0.0 and around 0.4 for the values chosen for $T$ and $S$ in Fig.~\ref{fig:phasediagramdeltaepsilon}. The AllD equilibrium is always stable.

\paragraph{Learnability.}
In this section, we show how the learning rate $\alpha$ affects the learnability of the robustly cooperative WSLS equilibrium. With learnability, we mean the likelihood that the learning process reaches an equilibrium, i.e., the size of the state-action-value space from which the WSLS is learned. Following the edges along the \textit{Mutual Best-Response Networks} represent the deterministic dynamics of a reinforcement learning algorithm, learning with perfect information and a learning rate of $\alpha=1$ \cite{Meylahn2021}. The maximum learnability of the WSLS equilibrium under these dynamics, as given by its basin of attraction, over all possible parameters ($T, S, \epsilon$ and $\delta$) is $0.015625$ (see \textit{Methods}). Cooperation is thus not very likely in this case.

\begin{figure*}
	\centering
	\begin{subfigure}[b]{0.95\textwidth}
         \centering
         \includegraphics[width=0.32\linewidth]{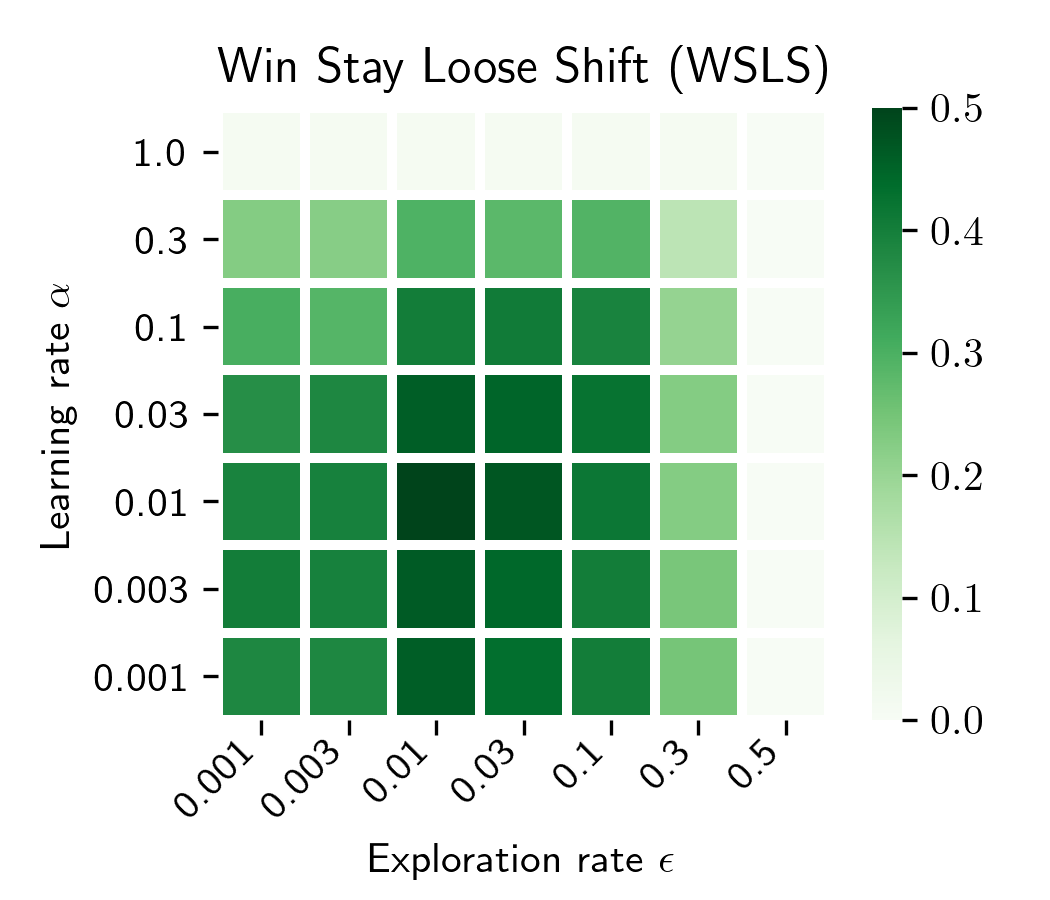}
         \includegraphics[width=0.32\linewidth]{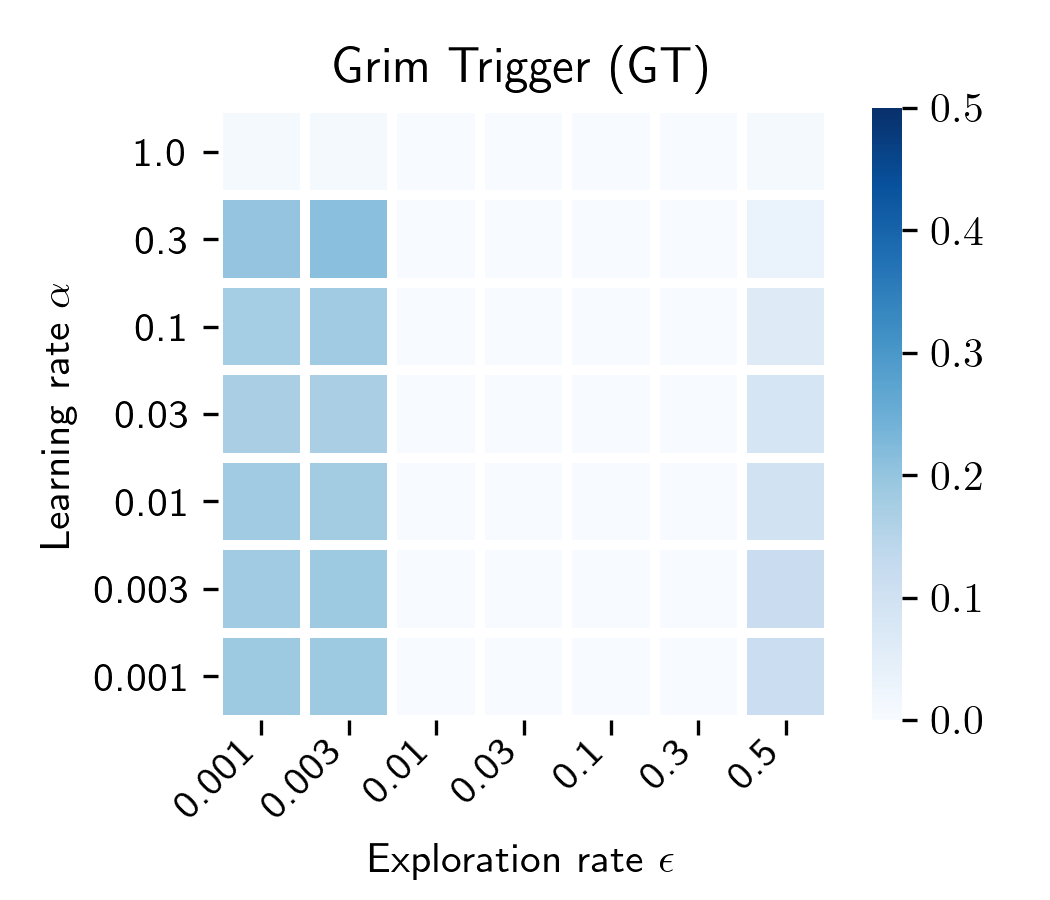}
         \includegraphics[width=0.32\linewidth]{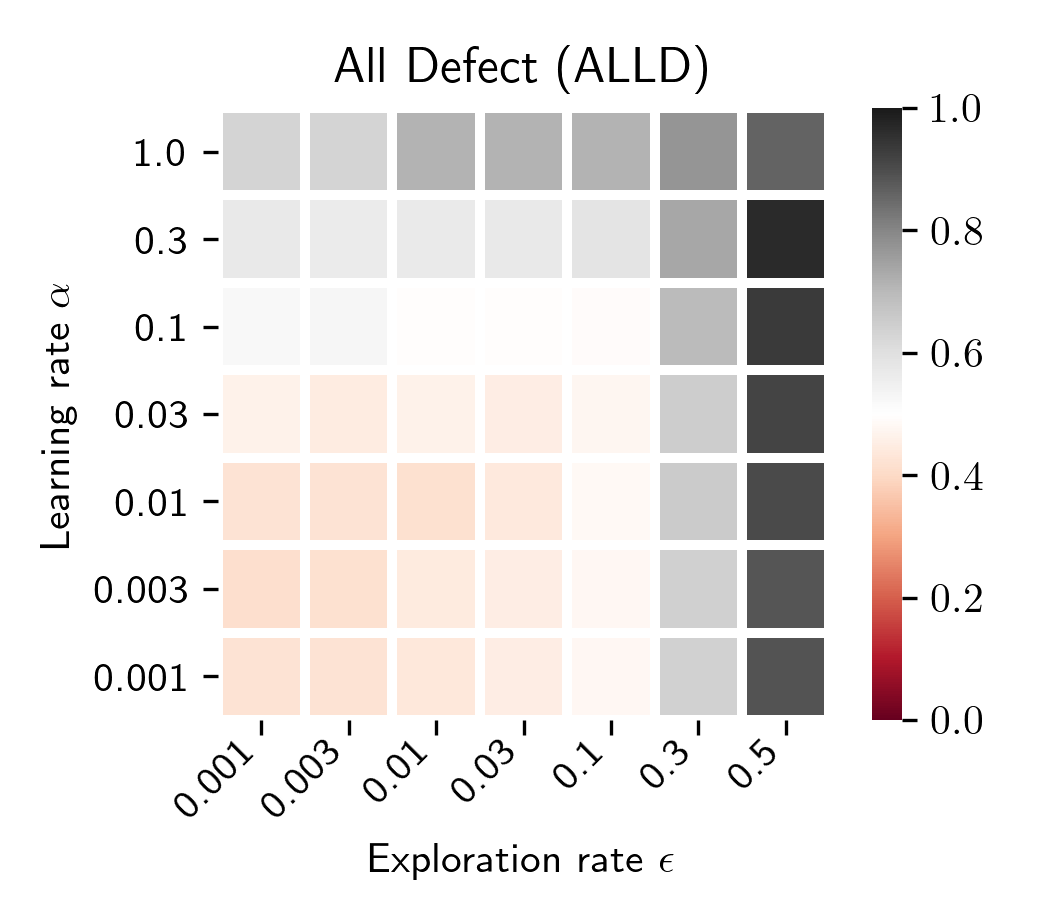}
         \caption{Environment parameters $S=-0.2, T=1.5$ }
     \end{subfigure}
     \hfill
		\begin{subfigure}[b]{0.95\textwidth}
         \centering
         \includegraphics[width=0.32\linewidth]{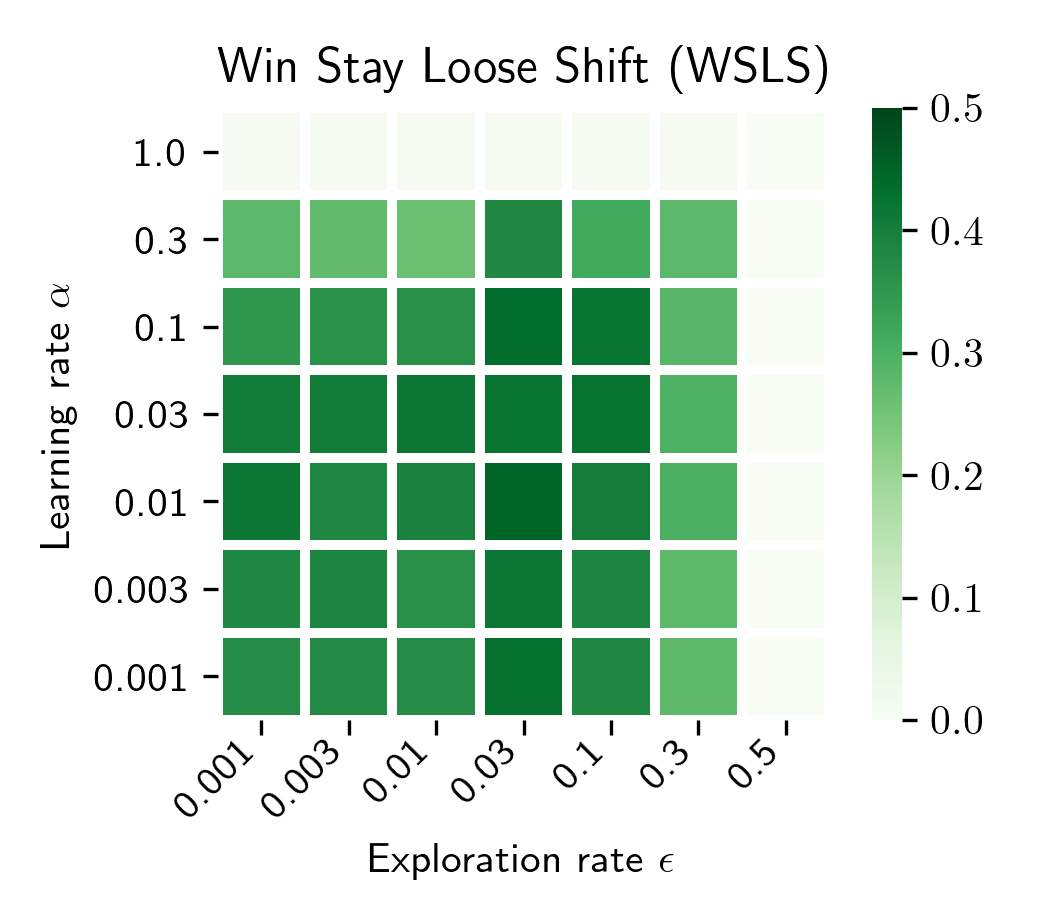}
         \includegraphics[width=0.32\linewidth]{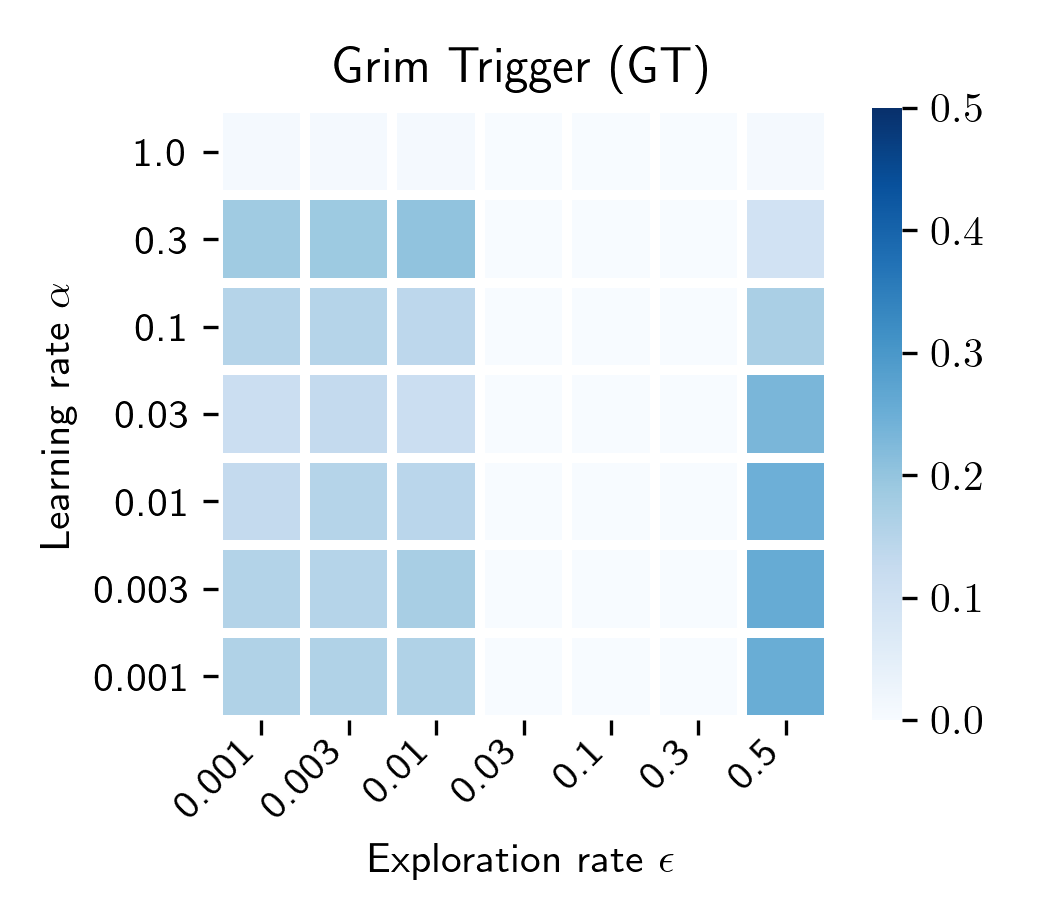}
         \includegraphics[width=0.32\linewidth]{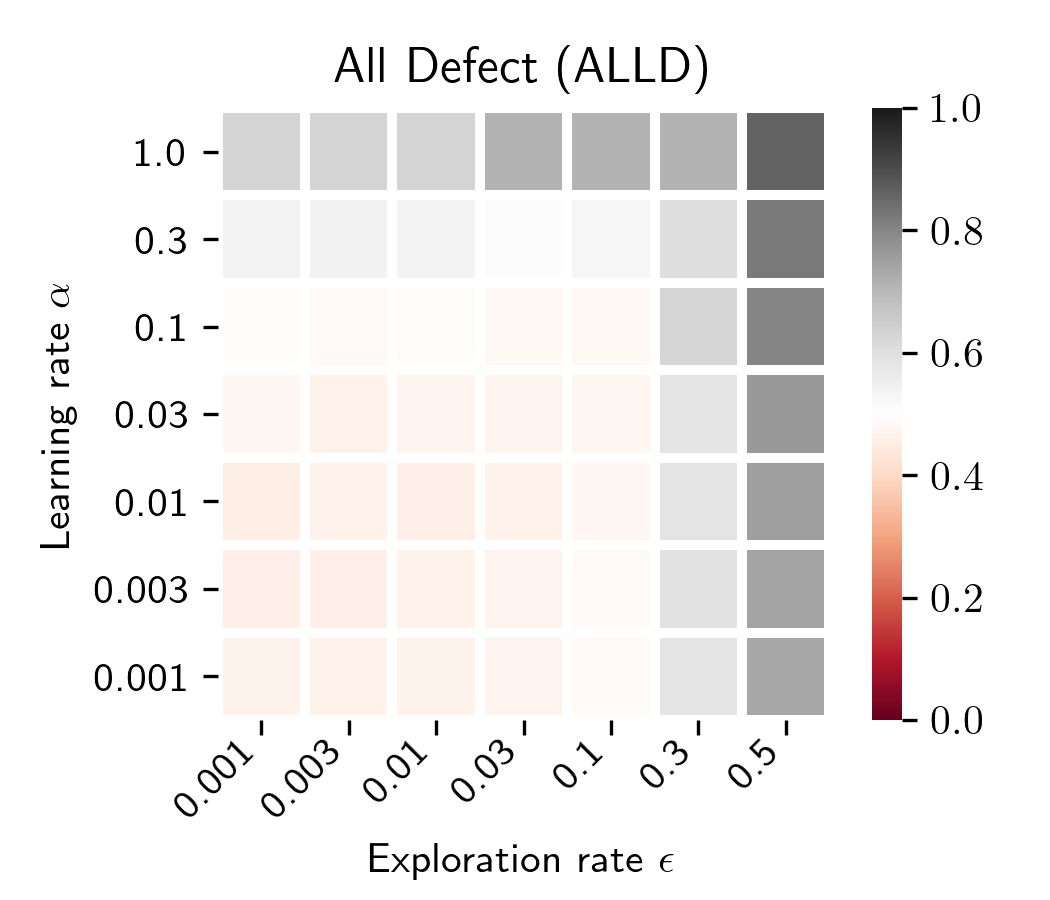}
         \caption{Environment parameters $S=-0.25, T=1.25$ }
     \end{subfigure}
     \hfill
	\caption{\textbf{Learnability parameter spaces.} Colors indicate which fraction of 250 random initial state-action values converges to the respective equilibria (the robustly cooperative WSLS in green on the left, GT in blue in the center, and AllD on the right) for a distinct parameter combination. Each plot portrays the parameter space spanned by the learning rate $\alpha$ versus the exploration rate $\epsilon$. The discount factor $\delta=0.99$. }
	\label{fig:learnability}
\end{figure*}    

To investigate the learning dynamics for a learning rate of $\alpha<1$, we refine the mathematical technique of deterministic reinforcement learning dynamics \cite{Barfuss2019,Barfuss2020,Barfuss2022}. This method considers the mean-field of an infinite memory batch to construct idealized learning updates precisely in the direction of the strategy-average temporal-difference error. Previous work investigated the dynamics in strategy space. We formulate the dynamics in state-action-value space to account for $\epsilon$-greedy policies (see \textit{Methods}).

To estimate the size of the state-action-value space from which the agents learn WSLS, we let them  start from  250 random initial state-action values (Fig.~\ref{fig:learnability}). Thus, with deterministic dynamics, the only randomness introduced in this section results from the initial state-action-value conditions.

Overall, we find that the robustly cooperative WSLS equilibrium is learned from a maximum of 40-50\% of the state-action-value space, given the learning rate $\alpha$ and the exploration rate $\epsilon$ are not too large (Fig.~\ref{fig:learnability}, left plots in green). Values below $0.1$  for each parameter are sufficient, independent of the environmental parameters investigated.

Furthermore, the deterministic learning dynamics confirm the non-trivial predictions of Fig.~\ref{fig:phasediagramdeltaepsilon} that the GT strategy is unstable for intermediate values of the exploration rates $\epsilon$ and large discount factors $\delta$ (Fig.~\ref{fig:learnability}, center plots in blue). Fig.~\ref{fig:learnability} also confirms the prediction that at exploration rates $\epsilon$ close to zero, the GT stability boundary is steeper for the environment with $S=-0.2$ and $T=1.5$ than for the environment with $S=-0.25$ and $T=1.25$ (Fig.~\ref{fig:phasediagramdeltaepsilon}). Grim Trigger is learned for exploration rate $\epsilon=0.01$ in the latter environment, but not in the former (Fig.~\ref{fig:learnability}).

Lastly, we find that outside the square of learning rate $\alpha=0.1$ and exploration rate $\epsilon=0.1$, more than half of the state-action-value space leads to the AllD equilibrium, independent of the environments investigated (Fig.~\ref{fig:learnability}, right plots). Inside this square, the AllD equilibrium no longer dominates the state-action-value space. Less than half of it leads to complete defection.

\paragraph{Stochasticity.}
In this section, we show that intrinsic fluctuations of the typical online reinforcement learning process significantly improve the learnability of the robustly cooperative WSLS equilibrium.
To be able to interpolate between fully online learning and deterministic learning, we refine the temporal-difference reinforcement learning algorithm (Eq.~\ref{eq:Qupdate}) with a memory batch of size $K \in \mathbb N$. 
\added{Batch learning is a prominent algorithmic refinement because of its efficient use of collected data and the improved stability of the learning process when used with function approximation \cite{lange2012batch,Lin1992, MnihEtAl2015}.}
The agents store experiences (observed states, rewards, next states) of $K$ time steps inside the memory batch and use their averages to get a more robust learning update of the state-action values (see \text{Methods}). The batch size $K$ allows us to interpolate between the fully online learning algorithm (Eq.~\ref{eq:Qupdate}) for $K=1$ and the deterministic learning dynamics for $K=\infty$. We simulate the stochastic batch-learning algorithm for an exemplary set of parameters to showcase the effect intrinsic fluctuations can have on the learning of cooperation. Our goal here is not to optimize this set of parameters, as a thorough theoretical treatment of the resulting stochastic process is beyond the scope of this work. 

\begin{figure*}
	\centering
    \begin{subfigure}[b]{0.48\textwidth}
        \centering
    	\includegraphics[scale=0.78]{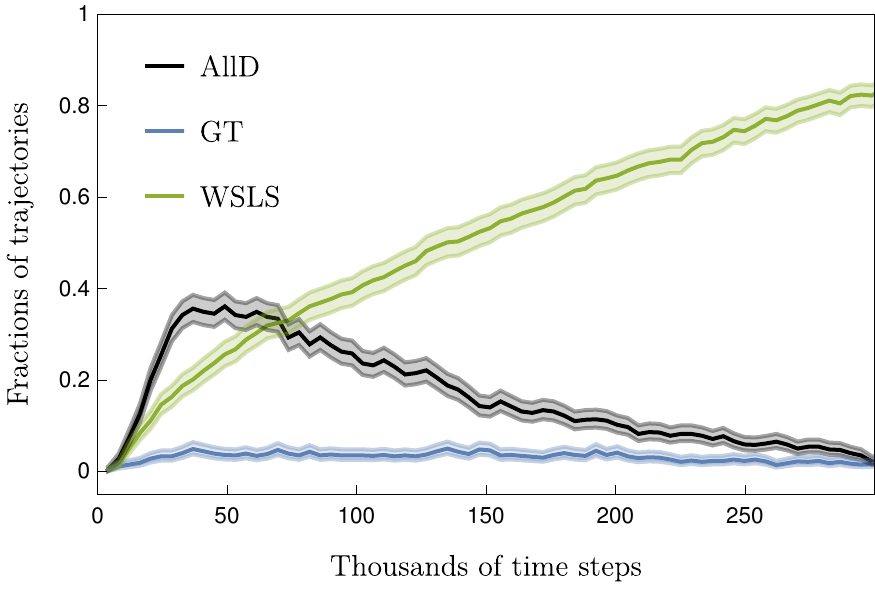}
		\vspace{-0.15cm}
    	\caption{Environment parameters $S=-0.2, T=1.5$ }
	\end{subfigure}
    \begin{subfigure}[b]{0.48\textwidth}
        \centering
	    \includegraphics[scale=0.78]{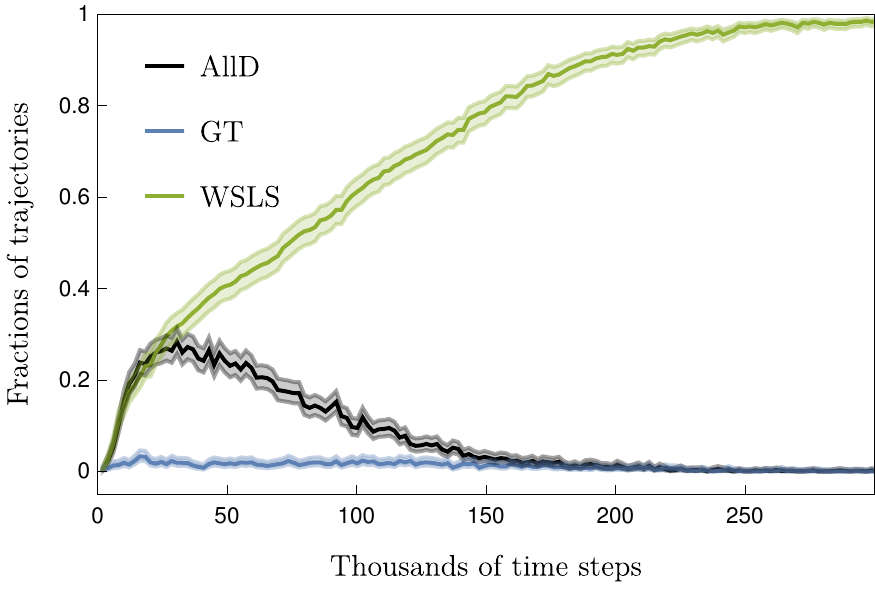}
		\vspace{-0.15cm}
        \caption{Environment parameters $S=-0.25, T=1.25$ }
	\end{subfigure}
	\caption{\textbf{Stochasticity of batch learning}. Fraction of 1000 from random initial state-action values in each of the three equilibria versus time steps for the sample-batch learning algorithm  with $\epsilon=0.1$, $\delta=0.99, \alpha=0.3$ and $K=4096$ for (a) and $K=2048$ for (b). The shaded region shows a 95\% confidence interval calculated using the Wilson Score  \cite{wilson1927probable}. }
	\label{fig:BatchTraj}
	\vspace{-0.15cm}
\end{figure*}

We find that intrinsic fluctuations significantly increase the level of the robustly cooperative WSLS equilibrium compared to the deterministic learning dynamics. At the same time, the batch learning agents require an order of magnitude fewer time steps to reach such high levels of cooperation than the batch-less online algorithm (Fig.~\ref{fig:BatchTraj}). We observe that the fraction of the cooperative equilibrium steadily increases to over 80\%, about twice the level reached with the deterministic learning dynamics in Fig~\ref{fig:learnability}. We hypothesize that the stability of the equilibria under noisy dynamics is a crucial factor. From Fig.~\ref{fig:BatchTraj}, we see that the percentage of trajectories in the WSLS state increases. In contrast, for the two other strategies, the percentage first increases and then decreases (with occasional upward fluctuations). This suggests that the WSLS strategy pair is more stable than the other two, given this choice of environment and algorithm parameters.

Interestingly, Fig.~\ref{fig:BatchTraj} also shows that the high level of robust cooperation is reached on a time scale that is an order of magnitude shorter than that of the purely online algorithm (Eq.~\ref{eq:Qupdate}). Whereas the online algorithm takes in the order of $10^6$ time steps, the batch learning algorithm only requires the order of $10^5$ time steps to reach high cooperation levels. This is remarkable because, in the batch-learning simulation, we purposefully restrict the agents to update their strategies only after completing an entire batch. In practice, learning a strategy using a memory batch and learning a model of the environment will be more intertwined \cite{VanSeijenSutton2015}, offering additional efficiency gains. This suggests the existence of a sweet spot between high levels of final cooperation and the  time agents require to learn them.

\added{To check the robustness of our results, we repeat the simulations of Figure \ref{fig:BatchTraj} for different combinations of the algorithm parameters $K, \alpha$ and $\epsilon$. The values we investigate are an increasing batch size $K\in\{1000, 2000, 3000, 4000, 5000, 6000, 7000, 8000, 9000, 10000\}$, for learning and exploration rate around the critical values $0.1$ which are decisive for high levels of robust cooperation in the deterministic approximation (Fig.~\ref{fig:learnability}): learning rate $\alpha \in \{0.003, 0.006, 0.1, 0.2, 0.3\}$ and exploration rate $\epsilon \in \{0.003, 0.006, 0.1, 0.2, 0.3\}$. We record the fraction of trajectories (based on 1000 samples) in the WSLS strategy pair at time $2\times 10^{6}$. To get an indication of the speed at which the WSLS strategy pair is learned, we record the time at which the fraction of trajectories for a given set of parameters reaches $0.4$. } 

\begin{figure}
    \centering
    \begin{subfigure}[b]{0.98\textwidth}
         \centering
         \includegraphics[width=0.98\linewidth]{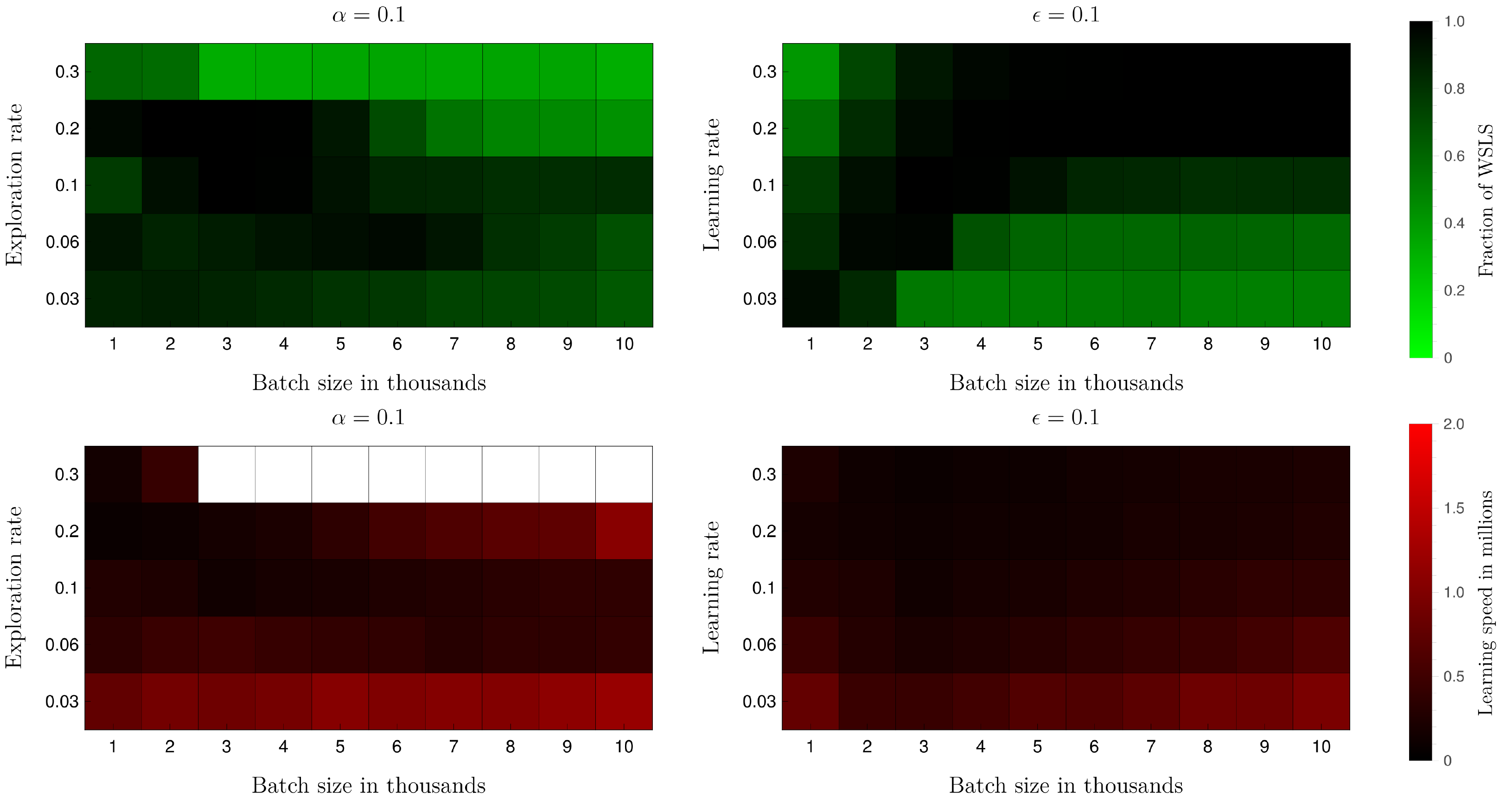}
         \caption{Environment parameters $S=-0.2, T=1.5$ }
     \end{subfigure}
     \hfill
		\begin{subfigure}[b]{0.98\textwidth}
         \centering
         \includegraphics[width=0.98\linewidth]{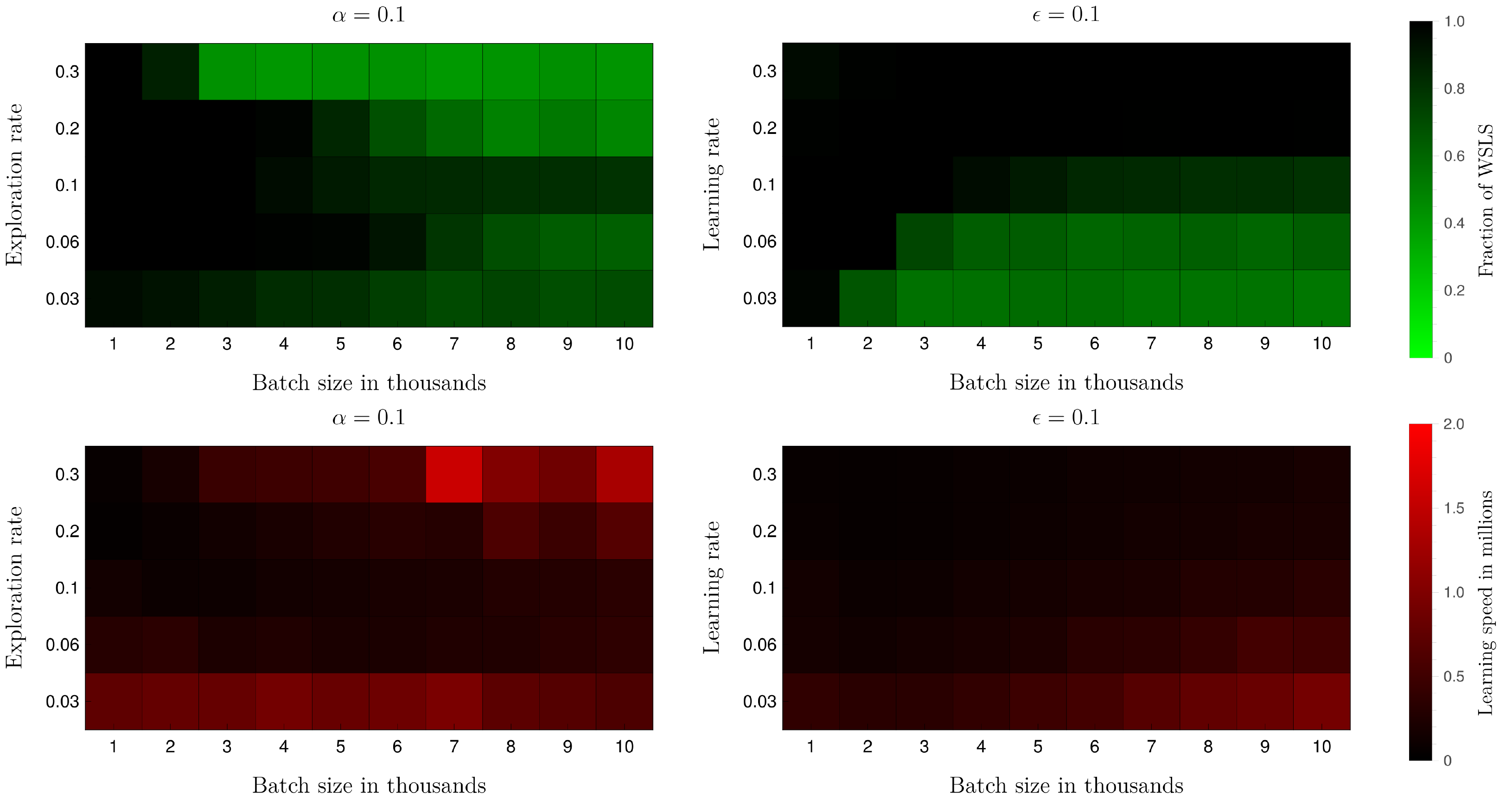}
         \caption{Environment parameters $S=-0.25, T=1.25$ }
     \end{subfigure}
     \hfill
    \caption{\added{\textbf{Robustness analysis for stochastic learning}. The green plots show the fraction of trajectories (1000 samples) that end in the WSLS strategy pair at time $2\times 10^{6}$, and the red plots show the time it takes for the fraction of trajectories in the WSLS strategy pair to reach $0.4$ in millions of time steps (we use white to represent trajectories that never reached $0.4$). The x-axis always represents the batch size in thousands, and the y-axis represents either the learning rate $\alpha$ or the exploration rate $\epsilon$. In all cases, we set $\delta=0.99$.}}
    \label{fig:robustness_main}
\end{figure}

\added{In Figure \ref{fig:robustness_main}, we show the results for both environments around the critical parameter space point $(\epsilon, \alpha)=(0.1, 0.1)$. The rest of the results are presented in the \textit{Supplementary Information}. We find that the fraction of trajectories in the WSLS strategy pair reaches values close to one for a large proportion of the investigated parameters. The results are thus robust to changes in the algorithm parameters. }

\added{Our robustness analysis also shows that intrinsic fluctuations do not make the other parameters irrelevant. We are able 
to draw some elementary conclusions regarding the combinations of parameters that lead to high levels of cooperation: (1) agents must not explore too much. Using an $\epsilon = 0.3$ leads to low levels of cooperation. (2) agents must not explore too little. We see that an $\epsilon = 0.03$ consistently leads to slower learning speeds than using intermediate values of the exploration rate. (3) larger learning rates lead to quicker learning speeds in the range of values we tested. In some cases, however, increasing the learning rate leads to lower levels of cooperation. The effect of changes in the batch size does not reveal a consistent pattern across the parameter ranges we tested. But if we restrict ourselves to the parameter values for the learning and exploration rates suggested by points (1)--(3), for example, $\alpha, \epsilon \in \{0.1, 0.2\}$, we see that an intermediate batch size of $K\in \{3000, 4000, 5000\}$ gives high levels of cooperation and achieves these quickly. Clearly, the interaction between these three parameters in how they influence the level of cooperation and the learning speed is complex. We leave a more detailed (theoretical) analysis of this interaction for future work.}

\section*{Discussion}
\label{sec:conclusion}

\paragraph{Contributions.}
In this article, we have shown that learning with imperfect\added{, inherently noisy} information is critical for the emergence of cooperation. We have done so by dissecting the widely used temporal-difference reinforcement learning process into three components. 

First,  cooperation can only be learned if a \textit{stable} equilibrium supports it. We have shown how the existence of all possible equilibria depends on the combination of environmental parameters, $T$, $S$, the agents' exploration rate, $\epsilon$, and how much they care for future rewards, $\delta$; under the assumption that the reinforcement learning update takes into account perfect information about the environment and the other agent's current strategy. The robustly cooperative Win-Stay, Lose-Shift (WSLS) equilibrium requires a small exploration rate, $\epsilon$, and a large discount factor $\delta$ to be stable (Fig.~\ref{fig:phasediagramdeltaepsilon}), but it is not the only stable equilibrium.

Second, cooperation will only be \textit{learned} if the WSLS equilibrium gets selected. This is more likely, the greater the size of the region of attraction leading to the WSLS equilibrium under the learning process.
We have shown for a large discount factor $\delta$ how the likelihood of learning all possible equilibria depends on the combination of the agents' learning and exploration rates $\alpha$ and $\epsilon$; as well as under the assumption that the reinforcement learning update takes into account perfect information about the environment and the other agent's current strategy. The robustly cooperative Win-Stay, Lose-Shift (WSLS) equilibrium requires a small $\alpha<0.1$ and a small $\epsilon<0.1$ to achieve cooperation levels of 40-50\% (Fig.~\ref{fig:learnability}). It is already interesting to observe that even though we give the reinforcement learners perfect information about the environment and the other agent's strategy, not using all of it for a learning update ($\alpha < 1$) is required to achieve cooperation levels of 40-50\%.

Third, we have shown that the internal \textit{stochasticity} of the learning process significantly improves the learnability of the robustly cooperative WSLS equilibrium. We have done so by simulating a sample batch version of the algorithm. Surprisingly, this algorithm learns to cooperate on a significantly shorter time scale than the online algorithm (Fig~\ref{fig:BatchTraj}). This highlights an essential trade-off between the cooperative learning outcome and the time it takes the agents to learn this outcome.
\added{For example, our finding suggests that in the seminal work of Sandholm and Crites \cite{SandholmCrites1996} the number of iterations and the amount of exploration for each trial was set too small to observe a high cooperation rate between two learning agents.}

The fact that intrinsic fluctuations of reinforcement learning promote cooperation is remarkable if we consider learning as a necessary tool to approximate an optimal solution when we don't have all information about the environment available. Indeed, temporal-difference learning will always converge to an optimal solution, given a decreasing learning rate and sufficient exploration \cite{sutton2018reinforcement}. However, this is only true in a single-agent environment. There, learning serves as a means to overcome a lack of information for optimal decision-making. More information could only improve learning and decision-making.

For multi-agent learning, the situation is radically different. We have shown that learning with imperfect information is not a necessary evil to overcome a lack of knowledge about the environment. Intrinsic fluctuations in the learning process are a crucial asset to learning collectively high-rewarding, cooperative solutions.
\deleted{This result is in general agreement that noise is not negligible in biological systems 
\cite{Bialek2012}, especially when it comes to the emergence of cooperation
\cite{NowakEtAl2004,JiaEtAl2010,AssafEtAl2013,Galla2009,Galla2011}.}

Methodologically, obtaining our result was possible by two complementary tools for studying strategy-average reinforcement learning dynamics in stylized games. We introduced mutual best-response networks for describing the dynamics in the strategy space and strategy-average learning dynamics for describing the learning in the value space. These methods are not tailored to investigate the iterated Prisoner's Dilemma. They are likewise applicable to derive insights from many other possible learning environments.

\paragraph{\added{Related work.}}
\added{
Our main result, that intrinsic fluctuations in temporal-difference reinforcement learning promote cooperation, is in general agreement with the result that noise in biological systems is not negligible \cite{Bialek2012}. With respect to evolutionary and learning dynamics, it is important to distinguish different noise concepts. 
}

\added{
Firstly, there is noise arising from suboptimal decision-making. In evolutionary game theory, such noise models the irrational or erroneous decision of players when adopting a less promising or rejecting a more promising strategy of another player. Such noise can be beneficial for cooperation \cite{VukovEtAl2006,SzolnokiEtAl2009,PercEtAl2017}. 
For individual learning, the analogous noise concept arises from the need to deviate from the currently optimal course of action to further explore the environment and improve the current strategy.
Thus, it is not necessarily irrational or erroneous to do so, but required for an individual learner. Analogous to evolutionary dynamics \cite{VukovEtAl2006}, this exploration parameter can cause bifurcations towards highly desirable equilibria \cite{LeonardosPiliouras2021}.
In our setting, the exploration rate, $\epsilon$, regulates this exploration-exploitation trade-off, and we show analytically that a small $\epsilon$ is required for robust cooperation (Fig.~\ref{fig:phasediagramdeltaepsilon}).
}

\added{
Secondly, external noise affecting the payoffs or rewards the agents receive can enhance cooperation in evolutionary dynamics \cite{JiaEtAl2010,AssafEtAl2013}. Similarly, it was recently shown that external L\'evy noise promotes cooperation \cite{WangEtAl2022} in reinforcement learning.
}

\added{
Thirdly, noise in the perceptions of agents can affect cooperation in learning and evolutionary dynamics \cite{SantosEtAl2021}. For example, inaccurate observations can lead to better learning outcomes in faster learning time, the stabilization of an otherwise chaotic learning process, and the mitigation of social dilemmas \cite{BarfussMann2022}.
}

\added{
Fourthly, there is the intrinsic noise of the evolutionary or learning process itself. 
In evolutionary game theory, such intrinsic noise arises because of finite populations, which can be highly beneficial for the evolution of cooperation \cite{NowakEtAl2004}.
With respect to learning dynamics, such intrinsic noise has been found to lead to noise-sustained cycling between cooperation and defection \cite{Galla2009,Galla2011,BladonGalla2011}.
This is the noise concept we are referring to when we speak about intrinsic fluctuations, and we have shown empirically how these fluctuations can be highly beneficial for the learning of cooperation.
}

\paragraph{Limitations and Future Work.} Our results show that understanding the effects of  intrinsic fluctuations in reinforcement learning is crucial in multi-agent systems. A formal treatment of these fluctuations is currently lacking and is an important avenue for future work. 

The time scale on which the agents learn cooperation  in our simulation with the sample batch algorithm is an order of magnitude faster than the online algorithm. Tuning the sample-batch-algorithm parameters, refining the algorithm with techniques such as optimism and leniency \cite{PanaitEtAl2006,PanaitEtAl2008}, and using more refined model-based variants \cite{VanSeijenSutton2015} may further improve the learning speed.

Our work has focused on $\epsilon$-greedy learning policies, which differ significantly from softmax exploration. Studying the learning dynamics under such policies will determine whether the results are a feature of exploration in general or are specific to $\epsilon$-greedy exploration. 

The environment of the iterated prisoner's dilemma is paradigmatic, but certainly not the only environment for studying cooperation. Our methods lend themselves to be applied in a variety of settings, such as a pricing duopoly with a discrete price space \cite{Calvano2020}, public goods games or common-pool resource harvesting with more than two learning agents \cite{BarfussEtAl2017,GeierEtAl2019}, and 
social dilemma situations with changing external environments \cite{BarfussEtAl2018,BarfussEtAl2020}.

\paragraph{Practical implications.}
Our results highlight that both designers of cooperative algorithms and regulators of algorithmic collusion must not focus solely on the learning outcome, but also on the learning efficiency. The existence of (online) algorithms that learn to cooperate under self-play is not sufficient for them to be applied in practice unless cooperation occurs on reasonable time scales, and they can learn reasonable strategies against a large class of algorithms currently employed in practice \cite{denBoer2022}. 

Overall, when designing sample batch algorithms, cooperation can be optimized, given the environment ($T$ and $S$), by choosing $\delta$, $\epsilon$, $\alpha$, and $K$ following three guiding criteria: (1) the cooperative equilibrium exists and has a relatively large basin of attraction, (2) the difference in stability between the cooperative equilibrium and the other equilibria is maximized in favor of the cooperative equilibrium, and that (3) the time scale on which cooperation is achieved is minimized.

\section*{Methods}

\paragraph{Mutual Best-Response Networks (MBRN).}
An $\epsilon$-greedy strategy can be characterized by a pure strategy, determined using the ordering of state-action values, together with exploration. If $\epsilon$ is fixed, all possible $\epsilon$-greedy strategies can be enumerated and represented using a four-dimensional vector. Given that the opponent plays a fixed $\epsilon$-greedy strategy, we can solve the Bellman equations to obtain the $\epsilon$-greedy strategy that is a best response. 

The state the system is in at any time, given two agents using an $\epsilon$-greedy strategy, is similarly characterized by an eight-dimensional vector representing both strategies. We refer to this eight-dimensional vector as a \textit{strategy pair}. A mutual best-response to a strategy pair is a strategy pair in which both agents play a strategy that is an $\epsilon$-greedy best response to the opponent's previous strategy. In this way, we construct a directed network (of 256 strategy pairs) with edges representing mutual best responses. 

By considering all possible strategy pairs, we can tabulate which edges are possible in the resulting MBRN, as well as the conditions for their existence. Taking the intersection of all possible combinations of edge conditions splits the parameter space into regions, so each region corresponds to a different MBRN (similar to the phase diagrams in \cite{Meylahn2021}). As a result, we can calculate maxima and minima over the entire parameter space by considering all possible MBRNs. 

The structure of this network depends on the reward parameters ($T$ and $S$), $\epsilon$, and $\delta$. Strategy pairs with self-loops are an equilibrium under mutual best-response dynamics. By solving the Bellman equations self-consistently, we can determine the critical conditions at which strategy pairs become equilibria. 

We define the fraction of strategy pairs that lead to an equilibrium under the mutual best-response dynamics as its basin of attraction. Given an initialization that selects an initial strategy pair uniformly at random from all possible strategy pairs, the basin of attraction of an equilibrium strategy pair represents the probability of ending in that strategy pair under the mutual best-response dynamics.

\paragraph{Learning Dynamics.}
In essence, deterministic temporal-difference learning dynamics use strategy averages instead of individual samples of obtained rewards and estimated next-state values. They model the idealized learning behavior of agents with an infinite memory batch \cite{Barfuss2020} or with separated time scales between the process of interaction and adaptation \cite{Barfuss2022}. Existing learning dynamic equations with $\epsilon$-greedy strategies were derived only for stateless interactions \cite{WunderEtAl2010}. State-full learning dynamics employed only softmax strategies \cite{Barfuss2019}. In the following, we present the deterministic Expected SARSA equations for state-full environments with $\epsilon$-greedy strategies in discrete time.

These dynamics operate in the joint state-action-value space $\tens q =\bigotimes_{i,s,a} q^{i}(s, a)$. In order to formulate the strategy-average update of $\tens q$ we define the joint strategy $\tens x =\bigotimes_{i,s,a} x^{i}(a|s) $ with $x^{i}(a|s)$ as the probability that agents $i$ will take action $a$ in state $s$. For $\epsilon$-greedy strategies, $\tens x$ is uniquely determined by $\tens q$ and $\epsilon$. To obtain deterministic dynamics, we need to derive the strategy-average version of the state-action update (Eq.~\ref{eq:Qupdate}),

\begin{align}
	\label{eq:detQupdate}
	q^{i}_{t+1}(s, a) = q^{i}_{t}(s, a)&+ \alpha\Big[r_{\tens{x}_t}^{i}(a|s) +\delta \cdot {}^{\text{next}}q_{\tens x_t}^{i}(s, a) - q^{i}_{t}(s, a)\Big],
\end{align}
where $r_{\tens{x}_t}^{i}(s, a)$ is the strategy-average version of the current reward and ${}^{\text{next}}q_{\tens{x}_t}^{i}(s, a)$ the strategy-average version of the expected value of the next state.

The strategy-average version of the current reward is obtained as
\begin{equation}
	r_{\tens{x}_t}^{i}(s, a) = \sum_{s'} \sum_{j \neq i} \prod_{a_j} x^{j}(a^j|s) \cdot  p(s'|\tens a, s) \cdot r^{i}(s').
\end{equation}
For each agent $i$, taking action $a$ in state $s$ when all other agents $j$ act according to their policies $x^{j}(a^j|s)$ causes the next state $s'$ via the \textit{transition probability} $p(s'|\tens a, s)$ at which agent $i$ obtains the reward, $r^{i}(s')$.

Second, the strategy-average version of the expected value of the next state is likewise computed by averaging over all actions of the other agents and next states. For each agent $i$ and state $s$, all other agents $j \neq i$ choose their action $a^j$ with probability $x^{j}(s,a^j)$. Consequently, the environment transitions to the next state $s'$ with probability $p(s'|\tens a, s)$. At $s'$, the agent estimates the quality to be the average of $q_{\tens{x}_t}^{i}(s',b)$ with respect to its own strategy. Mathematically, we write
\begin{align}
	{}^{\text{next}}q_{\tens{x}_t}^{i}(s, a) :=& \sum_{a^j} \sum_{s'} \prod_{j \neq i} x^{j}( a^j|s) p(s'|\tens a, s) \times\sum_{b}x^{i}(b|s') q_{\tens{x}_t}^{i}(s', b).
\end{align}
Here, we replace the quality estimates $q^{i}_{t}(s,a)$, which evolve in time $t$ (Eq.~\ref{eq:Qupdate}), with the strategy-average state-action quality, $q_{\tens{x}_t}^{i}(s, a)$, which is the expected discounted sum of future rewards from executing action $a$ in state $s$ and then following along the joint strategy $\tens x$.
It is obtained by adding the current strategy-average reward $r_{\tens{x}_t}^{i}(s,a)$ to the discounted strategy-average state quality of the next state $v_{\tens{x}_t}^{i}(s')$,
\begin{align}
	q_{\tens{x}_t}^{i}(s,a) = r_{\tens{x}_t}^{i}(s,a)
	+ \ \delta \sum_{s'} p_{\tens{x}_t}^{i}( s'| a^i, s) \cdot  v_{\tens{x}_t}^{i}(s').
	\label{eq:Qiha}
\end{align}
Here, $p_{\tens{x}_t}^{i}( s'| a^i, s)$ is agent $i$'s strategy-average transition probability to state $s'$ from state $s^i$ under action $a^i$.
It is computed by averaging over all actions of the other agents. For each agent $i$ at state $s$, selecting action $a^i$, all other agents $j \neq i$ select action $a^j$ with probability $x^{j}(a^j|s)$. Consequently, the environment will transition to the state $s'$ with probability $p(s'| \tens a, s)$. Mathematically, we write
\begin{align}
	p_{\tens{x}_t}^{i}( s'| a^i, s) =  \sum_{a^j} \prod_{j \neq i} x^{j}(a^j|s) \cdot p(s'| \tens a, s).
\end{align}

Further, at Eq.~\ref{eq:Qiha}, $v_{\tens{x}_t}^{i}(s)$ is the strategy-average state quality, i.e., the expected discounted sum of future rewards from state $s$ and then following along the joint strategy $\tens x$. They are computed via matrix inversion according to
\begin{equation}
	\tens v_{\tens{x}_t}^{i} = [
	\tens{\tens{\mathds{1}}}_{|\mathcal{S}|} -  \delta  \tens p_{\tens{x}_t} ]^{-1} \tens r_{\tens{x}_t}^{i},
	\label{eq:Vih}
\end{equation}
where $\tens v_{\tens{x}_t}^{i}$ denotes the $|\mathcal{S}|$-dimensional vector containing $v_{\tens{x}_t}^{i}(s)$ in entry $s$, $\tens r_{\tens{x}_t}^{i}$ is defined analogously and $\tens p_{\tens{x}_t}$ is a $|\mathcal{S}|\times |\mathcal{S}|$ matrix containing $p_{\tens{x}_t}(s, s')$ (defined in Eq.~\ref{eq:Tioo} below) at entry $(s, s')$.
Eq.~\ref{eq:Vih} is a direct conversion of the Bellman equation
$v_{\tens{x}_t}^{i}(s)= r_{\tens{x}_t}^{i}(s) + \delta \sum_{s'} p_{\tens{x}_t}(s, s') v_{\tens{x}_t}^{i}(s')$, which expresses that the value of the current observation is the discount factor weighted average of the current payoff and the value of the next state. Bold symbols indicate that the corresponding object is a vector or matrix, and $\tens{\tens{\mathds{1}}}_Z$ is the $Z$-by-$Z$ identity matrix.

The strategy-averaged transition matrix is denoted by $ \tens p_{\tens{x}_t}$. The entry $p_{\tens{x}_t}(s, s')$ indicates the probability that the environment will transition to state $s'$ after being in state $s$, given all agents follow the joint strategy $\tens x$. We compute them by averaging over all actions from all agents,
\begin{align}
	p_{\tens{x}_t}(s, s')= \sum_{a^j} \prod_{j} x^{j}(a|s) \cdot
	p(s'| \tens a, s).
	\label{eq:Tioo}
\end{align}

Further, in Eq.~\ref{eq:Vih}, $r_{\tens{x}_t}^{i}(s)$ denotes the strategy-average reward agent $i$ obtains at state $s$. We compute them by averaging all actions from all agents and all next states. For each $i$ at state $s$, all agents $j$ choose action $a^j$ with probability $x^{j}(a^j|s)$. Hence, the environment transitions to the next state $s$ with probability $p(s'|\tens a, s)$ and agent $i$ receives the reward $r(s')$,
\begin{align}
	r_{\tens{x}_t}^{i}(s) := \sum_{a^j} \prod_{j}
	x^{j}(a^j|s) p(s'| \tens a, s) r(s').
\end{align}
Note that the quality ${}^{\text{next}}q_{\tens{x}_t}^{i}(s, a)$ depends on $s$ and $a$ although it is the strategy-averaged expected value of the next state.

We finally obtained all necessary terms of state-full temporal-difference learning with $\epsilon$-greedy strategies in value space $\tens q$. Using an efficient python implementation, we can apply those learning equations for simulation studies to investigate multi-agent learning phenomena in a fast and deterministically reproducible way.

\paragraph{Batch Learning.}
The batch reinforcement learning problem was originally defined as learning the best strategy from a fixed set of a-priori-known transition samples \cite{lange2012batch}. However, our goal is to construct an algorithm able to interpolate between the fully online and fully deterministic version of the temporal-difference reinforcement learning process. 
The learning process is divided into two phases, an interaction phase, and an adaptation phase. During the interaction phase, the agent keeps its strategy fixed while interacting with its environment for $K$ timesteps, collecting state, action, and reward information. During the adaptation phase, the agent uses the collected information to update its strategy. Key is the use of two state-action-value tables, one for acting ($q_\text{act}$), the other for improved value estimation ($q_\text{val}$). While $q_\text{act}$ is kept constant during the interaction phase, $q_\text{val}$ is iteratively updated \cite{Barfuss2020, Barfuss2022}.

Furthermore, we use an auxiliary, time-dependent learning rate $\alpha(s, a, t_{s, a})$ for $q_\text{val}$ and a global learning rate $\alpha$ for $q_\text{act}$. Here $t_{s, a}$ is the local time of the state-action pair $(s, a)$, which is given by the number of times the state-action value $q_\text{val}(s, a)$ has been updated during the batch. Since the environment is kept fixed for the duration of the batch, each sample in the batch should be valued equally. This can be achieved by using a state, action, and time-dependent learning rate $\alpha(s, a, t)=\frac{1}{t+1}$ (Algorithm~\ref{alg:SampleBatch}). 

\begin{algorithm}
\caption{Sample-Batch Temporal-Difference Learning}\label{alg:SampleBatch}
Given learning rate $\alpha$, exploration rate $\epsilon$, discount factor $\delta$

\Begin{
    Initialize $q_{\text{act}}(s, a)=q_{\text{val}}(s, a)$ randomly.
    
    Initialize $p(s'|a, s), n(s, a)$ and $r(s, a)$ to zero.
    
    Set $x(a|s)$ as $\epsilon$-greedy strategy from $q_{\text{act}}(s, a)$.
    
    Observe current state $s$. 
    
    \Repeat{done}{
        \For{$k=1$ \textbf{to} $K$ }{\Comment{Interaction phase} 
        
        Execute action from $x(a|s)$;
        
        Observe reward $r$ and next state $s'$;
        
        Set $n(s, a)\leftarrow n(s, a) +1$;
        
        Set $p(s'| a, s)\leftarrow p(s'| a, s) + 1$;
        
        Set $r(s, a)\leftarrow r(s, a) +r$;
        
        Set $\tilde \alpha \leftarrow  \frac{1}{n(s, a)+1}$;
        
        Set $q_\text{val}(s, a) \leftarrow (1-\tilde\alpha)q_{\text{val}}(s, a) + \tilde \alpha \Big[r +\delta \sum_b x(b|s') q_\text{val}(s', b) \Big]$;
        
        Set $s\leftarrow s'$;
        }
        \ForEach{$\hat{s}, \hat{a},  \hat s'$}{\Comment{Adaption phase}
        
        Set $\tilde{r} \leftarrow \frac{r(\hat{s}, \hat{a})}{\max\{1, n(\hat{s}, \hat{a})\}}$;
        
        Set $\tilde v \leftarrow \sum_{b, z} \frac{p(z | \hat{s}, \hat{a})}{\max\{1, n(\hat{s}, \hat{a})\}} x(b|z) q_\text{val}(z, b)$;
        
        Set $q_{\text{act}}(\hat s, \hat a)\leftarrow (1-\alpha)q_{\text{act}}(\hat s, \hat a) + \alpha \big[\tilde r +\delta \tilde v \big]$;
        
        Set $x(\hat a|\hat s)$ as $\epsilon$-greedy strategy \big($q_{\text{act}}(\hat s, \hat a)$ \big);
        
        Set $q_{\text{val}}(\hat s, \hat a)\leftarrow q_{\text{act}}(\hat s, \hat a)$;
        
        Set $p(\hat s' | \hat a, \hat s), n(\hat s, \hat a),$ and $r(\hat s, \hat a)$ to zero;
        }
    }
}
\end{algorithm}

\subsection*{Data availability}
Code to reproduce all results is available at:\\ \url{https://github.com/wbarfuss/intrinsic-fluctuations-cooperation} and is archived at: \url{https://doi.org/10.5281/zenodo.7303593}.

\subsection*{Acknowledgements}
This work was supported by the German Federal Ministry of Education and Research (BMBF): Tuebingen AI Center, FKZ: 01IS18039A, and the Dutch Institute for Emergent Phenomena (DIEP) cluster at the University of Amsterdam.


\footnotesize
\setlength{\bibsep}{4pt plus 0.5ex}
\bibliography{bib}

\begin{thebibliography}{82}
\providecommand{\natexlab}[1]{#1}
\providecommand{\url}[1]{\texttt{#1}}
\expandafter\ifx\csname urlstyle\endcsname\relax
  \providecommand{\doi}[1]{doi: #1}\else
  \providecommand{\doi}{doi: \begingroup \urlstyle{rm}\Url}\fi

\bibitem[Dafoe et~al.(2021)Dafoe, Bachrach, Hadfield, Horvitz, Larson, and
  Graepel]{DafoeEtAl2021}
Allan Dafoe, Yoram Bachrach, Gillian Hadfield, Eric Horvitz, Kate Larson, and
  Thore Graepel.
\newblock Cooperative {{AI}}: Machines must learn to find common ground.
\newblock \emph{Nature}, 593\penalty0 (7857):\penalty0 33--36, 2021.
\newblock \doi{10.1038/d41586-021-01170-0}.
\newblock URL \url{https://www.nature.com/articles/d41586-021-01170-0}.

\bibitem[Bertino et~al.(2020)Bertino, {Doshi-Velez}, Gini, Lopresti, and
  Parkes]{BertinoEtAl2020}
Elisa Bertino, Finale {Doshi-Velez}, Maria Gini, Daniel Lopresti, and David
  Parkes.
\newblock Artificial {{Intelligence}} \& {{Cooperation}}.
\newblock Technical report, 2020.
\newblock URL
  \url{https://cra.org/ccc/resources/ccc-led-whitepapers/#2020-quadrennial-papers}.

\bibitem[Levin(2020)]{Levin2020}
Simon~A. Levin.
\newblock Collective {{Cooperation}}: {{From Ecological Communities}} to
  {{Global Governance}} and {{Back}}.
\newblock In \emph{Collective {{Cooperation}}: {{From Ecological Communities}}
  to {{Global Governance}} and {{Back}}}, pages 311--317. {Princeton University
  Press}, 2020.
\newblock ISBN 978-0-691-19532-2.
\newblock \doi{10.1515/9780691195322-025}.
\newblock URL
  \url{https://www.degruyter.com/document/doi/10.1515/9780691195322-025/html}.

\bibitem[Dawes(1980)]{Dawes1980}
R~M Dawes.
\newblock Social {{Dilemmas}}.
\newblock \emph{Annual Review of Psychology}, 31\penalty0 (1):\penalty0
  169--193, 1980.
\newblock \doi{10.1146/annurev.ps.31.020180.001125}.
\newblock URL \url{https://doi.org/10.1146/annurev.ps.31.020180.001125}.

\bibitem[Harrington(2018)]{Harrington2018}
Joseph~E Harrington.
\newblock Developing competition law for collusion by autonomous artificial
  agents.
\newblock \emph{Journal of Competition Law \& Economics}, 14\penalty0
  (3):\penalty0 331--363, 2018.
\newblock \doi{10.1093/joclec/nhz001}.
\newblock URL \url{https://doi.org/10.1093/joclec/nhz001}.

\bibitem[Axelrod and Hamilton(1981)]{AxelrodHamilton1981}
Robert Axelrod and William~D. Hamilton.
\newblock The {{Evolution}} of {{Cooperation}}.
\newblock \emph{Science}, 211\penalty0 (4489):\penalty0 1390--1396, 1981.
\newblock \doi{10.1126/science.7466396}.
\newblock URL \url{https://www.science.org/doi/abs/10.1126/science.7466396}.

\bibitem[Nowak and Sigmund(1993)]{NowakSigmund1993}
Martin Nowak and Karl Sigmund.
\newblock A strategy of win-stay, lose-shift that outperforms tit-for-tat in
  the {{Prisoner}}'s {{Dilemma}} game.
\newblock \emph{Nature}, 364\penalty0 (6432):\penalty0 56--58, 1993.
\newblock ISSN 1476-4687.
\newblock \doi{10.1038/364056a0}.
\newblock URL \url{https://www.nature.com/articles/364056a0}.

\bibitem[Nowak(2006)]{Nowak2006}
Martin~A. Nowak.
\newblock Five {{Rules}} for the {{Evolution}} of {{Cooperation}}.
\newblock \emph{Science}, 314\penalty0 (5805), 2006.
\newblock \doi{10.1126/science.1133755}.
\newblock URL \url{https://www.science.org/doi/abs/10.1126/science.1133755}.

\bibitem[Perc et~al.(2013)Perc, {G{\'o}mez-Gardenes}, Szolnoki, Flor{\'i}a, and
  Moreno]{PercEtAl2013}
Matja{\v z} Perc, Jes{\'u}s {G{\'o}mez-Gardenes}, Attila Szolnoki, Luis~M.
  Flor{\'i}a, and Yamir Moreno.
\newblock Evolutionary dynamics of group interactions on structured
  populations: A review.
\newblock \emph{Journal of the royal society interface}, 10\penalty0
  (80):\penalty0 20120997, 2013.
\newblock \doi{10.1098/rsif.2012.0997}.
\newblock URL \url{https://doi.org/10.1098/rsif.2012.0997}.

\bibitem[Perc et~al.(2017)Perc, Jordan, Rand, Wang, Boccaletti, and
  Szolnoki]{PercEtAl2017}
Matja{\v z} Perc, Jillian~J. Jordan, David~G. Rand, Zhen Wang, Stefano
  Boccaletti, and Attila Szolnoki.
\newblock Statistical physics of human cooperation.
\newblock \emph{Physics Reports}, 687:\penalty0 1--51, 2017.
\newblock ISSN 0370-1573.
\newblock \doi{10.1016/j.physrep.2017.05.004}.
\newblock URL
  \url{https://www.sciencedirect.com/science/article/pii/S0370157317301424}.

\bibitem[Masuda and Ohtsuki(2009)]{MasudaOhtsuki2009}
Naoki Masuda and Hisashi Ohtsuki.
\newblock A {{Theoretical Analysis}} of {{Temporal Difference Learning}}
  in~the~{{Iterated Prisoner}}'s {{Dilemma Game}}.
\newblock \emph{Bulletin of Mathematical Biology}, 71\penalty0 (8):\penalty0
  1818--1850, 2009.
\newblock ISSN 1522-9602.
\newblock \doi{10.1007/s11538-009-9424-8}.
\newblock URL \url{https://doi.org/10.1007/s11538-009-9424-8}.

\bibitem[Ezrachi and Stucke(2016)]{EzrachiStucke2016}
A.~Ezrachi and M.E. Stucke.
\newblock \emph{Virtual competition: the promise and perils of the
  algorithm-driven economy}.
\newblock Cambridge, Massachusetts : Harvard University Press, 2016.

\bibitem[Cimini and Sánchez(2014)]{CiminiSanchez2014}
Giulio Cimini and Angel Sánchez.
\newblock Learning dynamics explains human behaviour in {Prisoner}'s {Dilemma}
  on networks.
\newblock \emph{Journal of The Royal Society Interface}, 11\penalty0
  (94):\penalty0 20131186, May 2014.
\newblock \doi{10.1098/rsif.2013.1186}.
\newblock URL
  \url{https://royalsocietypublishing.org/doi/full/10.1098/rsif.2013.1186}.
\newblock Publisher: Royal Society.

\bibitem[Ezaki et~al.(2016)Ezaki, Horita, Takezawa, and Masuda]{EzakiEtAl2016}
Takahiro Ezaki, Yutaka Horita, Masanori Takezawa, and Naoki Masuda.
\newblock Reinforcement {Learning} {Explains} {Conditional} {Cooperation} and
  {Its} {Moody} {Cousin}.
\newblock \emph{PLOS Computational Biology}, 12\penalty0 (7):\penalty0
  e1005034, July 2016.
\newblock ISSN 1553-7358.
\newblock \doi{10.1371/journal.pcbi.1005034}.
\newblock URL \url{https://dx.plos.org/10.1371/journal.pcbi.1005034}.

\bibitem[Ezrachi and Stucke(2017)]{EzrachiStucke2017}
Ariel Ezrachi and Maurice~E Stucke.
\newblock Artificial intelligence \& collusion: When computers inhibit
  competition.
\newblock \emph{U. Ill. L. Rev.}, page 1775, 2017.

\bibitem[Perolat et~al.(2017)Perolat, Leibo, Zambaldi, Beattie, Tuyls, and
  Graepel]{PerolatEtAl2017}
Julien Perolat, Joel~Z. Leibo, Vinicius Zambaldi, Charles Beattie, Karl Tuyls,
  and Thore Graepel.
\newblock A multi-agent reinforcement learning model of common-pool resource
  appropriation.
\newblock In \emph{Proceedings of the 31st {{International Conference}} on
  {{Neural Information Processing Systems}}}, {{NIPS}}'17, pages 3646--3655,
  {Red Hook, NY, USA}, 2017. {Curran Associates Inc.}
\newblock ISBN 978-1-5108-6096-4.
\newblock URL
  \url{https://proceedings.neurips.cc/paper/2017/file/2b0f658cbffd284984fb11d90254081f-Paper.pdf}.

\bibitem[Leibo et~al.(2017)Leibo, Zambaldi, Lanctot, Marecki, and
  Graepel]{LeiboEtAl2017}
Joel~Z. Leibo, Vinicius Zambaldi, Marc Lanctot, Janusz Marecki, and Thore
  Graepel.
\newblock Multi-agent {{Reinforcement Learning}} in {{Sequential Social
  Dilemmas}}.
\newblock In \emph{Proceedings of the 16th {{Conference}} on {{Autonomous
  Agents}} and {{MultiAgent Systems}}}, {{AAMAS}} '17, pages 464--473,
  {Richland, SC}, 2017. {International Foundation for Autonomous Agents and
  Multiagent Systems}.

\bibitem[K\"uhn and Tadelis(2017)]{KuhnTadelis2017}
K.-U. K\"uhn and S.~Tadelis.
\newblock Algorithmic collusion.
\newblock Presentation prepared for CRESSE, 2017.

\bibitem[Calvano et~al.(2019)Calvano, Calzolari, Denicol{\`o}, and
  Pastorello]{Calvano2019a}
Emilio Calvano, Giacomo Calzolari, Vincenzo Denicol{\`o}, and Sergio
  Pastorello.
\newblock Algorithmic pricing what implications for competition policy?
\newblock \emph{Review of industrial organization}, 55\penalty0 (1):\penalty0
  155--171, 2019.
\newblock \doi{10.1007/s11151-019-09689-3}.
\newblock URL \url{https://doi.org/10.1007/s11151-019-09689-3}.

\bibitem[Barbosa et~al.(2020)Barbosa, Costa, Melo, Sichman, and
  Santos]{BarbosaEtAl2020}
Jo{\~a}o~Vitor Barbosa, Anna H~Reali Costa, Francisco~S Melo, Jaime~S Sichman,
  and Francisco~C Santos.
\newblock Emergence of {{Cooperation}} in {{N-Person Dilemmas}} through
  {{Actor-Critic Reinforcement Learning}}.
\newblock In \emph{Proc. of the {{Adaptive}} and {{Learning Agents Workshop}}
  ({{ALA}} 2020),}, page~9, 2020.

\bibitem[Sandholm and Crites(1996)]{SandholmCrites1996}
Tuomas~W. Sandholm and Robert~H. Crites.
\newblock Multiagent reinforcement learning in the {{Iterated Prisoner}}'s
  {{Dilemma}}.
\newblock \emph{Biosystems}, 37\penalty0 (1):\penalty0 147--166, 1996.
\newblock ISSN 0303-2647.
\newblock \doi{10.1016/0303-2647(95)01551-5}.
\newblock URL
  \url{https://www.sciencedirect.com/science/article/pii/0303264795015515}.

\bibitem[Schrepel(2017)]{Schrepel2017}
T.~Schrepel.
\newblock Here's why algorithms are {NOT} (really) a thing.
\newblock \emph{Concurrentialiste}, May 2017 (online), 2017.

\bibitem[Schwalbe(2018)]{Schwalbe2018}
Ulrich Schwalbe.
\newblock Algorithms, machine learning, and collusion.
\newblock \emph{Journal of Competition Law {\&} Economics}, 14\penalty0
  (4):\penalty0 568--607, 2018.
\newblock \doi{10.1093/joclec/nhz004}.
\newblock URL \url{https://doi.org/10.1093/joclec/nhz004}.

\bibitem[Peysakhovich and Lerer(2018{\natexlab{a}})]{PeysakhovichLerer2018}
Alexander Peysakhovich and Adam Lerer.
\newblock Towards {{AI}} that can solve social dilemmas.
\newblock In \emph{{{AAAI Spring Symposium Series}}}, page~7,
  2018{\natexlab{a}}.

\bibitem[Dafoe et~al.(2020)Dafoe, Hughes, Bachrach, Collins, McKee, Leibo,
  Larson, and Graepel]{DafoeEtAl2020}
Allan Dafoe, Edward Hughes, Yoram Bachrach, Tantum Collins, Kevin~R. McKee,
  Joel~Z. Leibo, Kate Larson, and Thore Graepel.
\newblock Open {{Problems}} in {{Cooperative AI}}.
\newblock \emph{arXiv preprint}, 2020.
\newblock URL \url{https://arxiv.org/abs/2012.08630v1}.

\bibitem[Peysakhovich and Lerer(2018{\natexlab{b}})]{PeysakhovichLerer2018a}
Alexander Peysakhovich and Adam Lerer.
\newblock Consequentialist conditional cooperation in social dilemmas with
  imperfect information.
\newblock In \emph{International {{Conference}} on {{Learning
  Representations}}}, 2018{\natexlab{b}}.
\newblock URL \url{https://openreview.net/forum?id=BkabRiQpb}.

\bibitem[Lerer and Peysakhovich(2018)]{LererPeysakhovich2018}
Adam Lerer and Alexander Peysakhovich.
\newblock Maintaining cooperation in complex social dilemmas using deep
  reinforcement learning.
\newblock 2018.
\newblock URL \url{https://arxiv.org/abs/1707.01068v4}.

\bibitem[Foerster et~al.(2018)Foerster, Chen, {Al-Shedivat}, Whiteson, Abbeel,
  and Mordatch]{FoersterEtAl2018}
Jakob Foerster, Richard~Y. Chen, Maruan {Al-Shedivat}, Shimon Whiteson, Pieter
  Abbeel, and Igor Mordatch.
\newblock Learning with {{Opponent-Learning Awareness}}.
\newblock In \emph{Proceedings of the 17th {{International Conference}} on
  {{Autonomous Agents}} and {{MultiAgent Systems}}}, {{AAMAS}} '18, pages
  122--130, {Richland, SC}, 2018. {International Foundation for Autonomous
  Agents and Multiagent Systems}.

\bibitem[Hughes et~al.(2018)Hughes, Leibo, Phillips, Tuyls,
  {Due{\~n}ez-Guzman}, Garc{\'i}a~Casta{\~n}eda, Dunning, Zhu, McKee, Koster,
  Roff, and Graepel]{HughesEtAl2018}
Edward Hughes, Joel~Z Leibo, Matthew Phillips, Karl Tuyls, Edgar
  {Due{\~n}ez-Guzman}, Antonio Garc{\'i}a~Casta{\~n}eda, Iain Dunning, Tina
  Zhu, Kevin McKee, Raphael Koster, Heather Roff, and Thore Graepel.
\newblock Inequity aversion improves cooperation in intertemporal social
  dilemmas.
\newblock In \emph{Advances in {{Neural Information Processing Systems}}},
  volume~31. {Curran Associates, Inc.}, 2018.
\newblock URL
  \url{https://proceedings.neurips.cc/paper/2018/hash/7fea637fd6d02b8f0adf6f7dc36aed93-Abstract.html}.

\bibitem[Eccles et~al.(2019)Eccles, Hughes, Kram{\'a}r, Wheelwright, and
  Leibo]{EcclesEtAl2019}
Tom Eccles, Edward Hughes, J{\'a}nos Kram{\'a}r, Steven Wheelwright, and Joel~Z
  Leibo.
\newblock The {{Imitation Game}}: {{Learned Reciprocity}} in {{Markov}} games.
\newblock In \emph{{{AAMAS}} '19: {{Proceedings}} of the 18th {{International
  Conference}} on {{Autonomous Agents}} and {{MultiAgent Systems}}}, page~3,
  2019.

\bibitem[Baker(2020)]{Baker2020}
Bowen Baker.
\newblock Emergent reciprocity and team formation from randomized uncertain
  social preferences.
\newblock In H.~Larochelle, M.~Ranzato, R.~Hadsell, M.~F. Balcan, and H.~Lin,
  editors, \emph{Advances in Neural Information Processing Systems}, volume~33,
  pages 15786--15799. {Curran Associates, Inc.}, 2020.
\newblock URL
  \url{https://proceedings.neurips.cc/paper/2020/file/b63c87b0a41016ad29313f0d7393cee8-Paper.pdf}.

\bibitem[Wang et~al.(2019)Wang, Hughes, Fernando, Czarnecki, Du\'{e}\~{n}ez
  Guzm\'{a}n, and Leibo]{WangEtAl2019}
Jane~X. Wang, Edward Hughes, Chrisantha Fernando, Wojciech~M. Czarnecki,
  Edgar~A. Du\'{e}\~{n}ez Guzm\'{a}n, and Joel~Z. Leibo.
\newblock Evolving intrinsic motivations for altruistic behavior.
\newblock In \emph{Proceedings of the 18th International Conference on
  Autonomous Agents and MultiAgent Systems}, AAMAS '19, page 683–692,
  Richland, SC, 2019. International Foundation for Autonomous Agents and
  Multiagent Systems.
\newblock ISBN 9781450363099.

\bibitem[Hughes et~al.(2020)Hughes, Anthony, Eccles, Leibo, Balduzzi, and
  Bachrach]{HughesEtAl2020}
Edward Hughes, Thomas~W. Anthony, Tom Eccles, Joel~Z. Leibo, David Balduzzi,
  and Yoram Bachrach.
\newblock Learning to resolve alliance dilemmas in many-player zero-sum games.
\newblock In \emph{Proceedings of the 19th International Conference on
  Autonomous Agents and MultiAgent Systems}, AAMAS '20, page 538–547,
  Richland, SC, 2020. International Foundation for Autonomous Agents and
  Multiagent Systems.
\newblock ISBN 9781450375184.

\bibitem[Meylahn and den Boer(2022)]{Meylahn2022}
Janusz~M Meylahn and Arnoud~V den Boer.
\newblock Learning to collude in a pricing duopoly.
\newblock \emph{Manufacturing \& Service Operations Management}, 24\penalty0
  (5), 2022.
\newblock \doi{10.1287/msom.2021.1074}.
\newblock URL \url{https://doi.org/10.1287/msom.2021.1074}.

\bibitem[Bowling and Veloso(2002)]{BowlingVeloso2002}
Michael Bowling and Manuela Veloso.
\newblock Multiagent learning using a variable learning rate.
\newblock \emph{Artificial Intelligence}, 136\penalty0 (2):\penalty0 215--250,
  2002.
\newblock ISSN 00043702.
\newblock \doi{10.1016/S0004-3702(02)00121-2}.
\newblock URL
  \url{https://linkinghub.elsevier.com/retrieve/pii/S0004370202001212}.

\bibitem[{de Cote} et~al.(2006){de Cote}, Lazaric, and
  Restelli]{deCoteEtAl2006}
Enrique~Munoz {de Cote}, Alessandro Lazaric, and Marcello Restelli.
\newblock Learning to cooperate in multi-agent social dilemmas.
\newblock In \emph{Proceedings of the Fifth International Joint Conference on
  {{Autonomous}} Agents and Multiagent Systems}, {{AAMAS}} '06, pages 783--785,
  {New York, NY, USA}, 2006. {Association for Computing Machinery}.
\newblock ISBN 978-1-59593-303-4.
\newblock \doi{10.1145/1160633.1160770}.
\newblock URL \url{https://doi.org/10.1145/1160633.1160770}.

\bibitem[Panait et~al.(2006)Panait, Sullivan, and Luke]{PanaitEtAl2006}
Liviu Panait, Keith Sullivan, and Sean Luke.
\newblock Lenient learners in cooperative multiagent systems.
\newblock In \emph{Proceedings of the Fifth International Joint Conference on
  {{Autonomous}} Agents and Multiagent Systems}, pages 801--803, 2006.

\bibitem[Stimpson and Goodrich(2003)]{StimpsonGoodrich2003}
Jeffrey~L. Stimpson and Michael~A. Goodrich.
\newblock Learning to cooperate in a social dilemma: A satisficing approach to
  bargaining.
\newblock In \emph{Proceedings of the {{Twentieth International Conference}} on
  {{International Conference}} on {{Machine Learning}}}, {{ICML}}'03, pages
  728--735, {Washington, DC, USA}, 2003. {AAAI Press}.
\newblock ISBN 978-1-57735-189-4.

\bibitem[Bush and Mosteller(1951)]{BushMosteller1951}
Robert~R. Bush and Frederick Mosteller.
\newblock A mathematical model for simple learning.
\newblock \emph{Psychological Review}, 58:\penalty0 313--323, 1951.
\newblock ISSN 1939-1471.
\newblock \doi{10.1037/h0054388}.
\newblock URL \url{https://psycnet.apa.org/doi/10.1037/h0054388}.

\bibitem[Macy and Flache(2002)]{MacyFlache2002}
Michael~W. Macy and Andreas Flache.
\newblock Learning dynamics in social dilemmas.
\newblock \emph{Proceedings of the National Academy of Sciences}, 99\penalty0
  (suppl\_3):\penalty0 7229--7236, May 2002.
\newblock \doi{10.1073/pnas.092080099}.
\newblock URL \url{https://www.pnas.org/doi/abs/10.1073/pnas.092080099}.
\newblock Publisher: Proceedings of the National Academy of Sciences.

\bibitem[Izquierdo et~al.(2008)Izquierdo, Izquierdo, and
  Gotts]{IzquierdoEtAl2008}
Segismundo~S. Izquierdo, Luis~R. Izquierdo, and Nicholas~M. Gotts.
\newblock Reinforcement learning dynamics in social dilemmas.
\newblock \emph{Journal of Artificial Societies and Social Simulation},
  11\penalty0 (2):\penalty0 1, 2008.
\newblock URL \url{https://www.jasss.org/11/2/1.html}.

\bibitem[Masuda and Nakamura(2011)]{MasudaNakamura2011}
Naoki Masuda and Mitsuhiro Nakamura.
\newblock Numerical analysis of a reinforcement learning model with the dynamic
  aspiration level in the iterated {{Prisoner}}'s dilemma.
\newblock \emph{Journal of Theoretical Biology}, 278\penalty0 (1):\penalty0
  55--62, 2011.
\newblock ISSN 0022-5193.
\newblock \doi{10.1016/j.jtbi.2011.03.005}.
\newblock URL
  \url{https://www.sciencedirect.com/science/article/pii/S0022519311001342}.

\bibitem[Zhang et~al.(2012)Zhang, Wu, and Wang]{ZhangEtAl2012}
Hai-Feng Zhang, Zhi-Xi Wu, and Bing-Hong Wang.
\newblock Universal effect of dynamical reinforcement learning mechanism in
  spatial evolutionary games.
\newblock \emph{Journal of Statistical Mechanics: Theory and Experiment},
  2012\penalty0 (06):\penalty0 P06005, 2012.
\newblock ISSN 1742-5468.
\newblock \doi{10.1088/1742-5468/2012/06/P06005}.
\newblock URL \url{https://dx.doi.org/10.1088/1742-5468/2012/06/P06005}.

\bibitem[Jia and Ma(2013)]{JiaMa2013}
Ning Jia and Shoufeng Ma.
\newblock Evolution of cooperation in the snowdrift game among mobile players
  with random-pairing and reinforcement learning.
\newblock \emph{Physica A: Statistical Mechanics and its Applications},
  392\penalty0 (22):\penalty0 5700--5710, 2013.
\newblock ISSN 0378-4371.
\newblock \doi{10.1016/j.physa.2013.07.049}.
\newblock URL
  \url{https://www.sciencedirect.com/science/article/pii/S0378437113006596}.

\bibitem[Jia et~al.(2021)Jia, Guo, Song, Shi, Deng, Perc, and
  Wang]{JiaEtAl2021}
Danyang Jia, Hao Guo, Zhao Song, Lei Shi, Xinyang Deng, Matja{\v z} Perc, and
  Zhen Wang.
\newblock Local and global stimuli in reinforcement learning.
\newblock \emph{New Journal of Physics}, 23\penalty0 (8):\penalty0 083020,
  2021.
\newblock ISSN 1367-2630.
\newblock \doi{10.1088/1367-2630/ac170a}.
\newblock URL \url{https://dx.doi.org/10.1088/1367-2630/ac170a}.

\bibitem[Song et~al.(2022)Song, Guo, Jia, Perc, Li, and Wang]{SongEtAl2022}
Zhao Song, Hao Guo, Danyang Jia, Matja{\v z} Perc, Xuelong Li, and Zhen Wang.
\newblock Reinforcement learning facilitates an optimal interaction intensity
  for cooperation.
\newblock \emph{Neurocomputing}, 513:\penalty0 104--113, 2022.
\newblock ISSN 0925-2312.
\newblock \doi{10.1016/j.neucom.2022.09.109}.
\newblock URL
  \url{https://www.sciencedirect.com/science/article/pii/S0925231222012000}.

\bibitem[Botvinick et~al.(2020)Botvinick, Wang, Dabney, Miller, and
  {Kurth-Nelson}]{BotvinickEtAl2020}
Matthew Botvinick, Jane~X. Wang, Will Dabney, Kevin~J. Miller, and Zeb
  {Kurth-Nelson}.
\newblock Deep {{Reinforcement Learning}} and {{Its Neuroscientific
  Implications}}.
\newblock \emph{Neuron}, 107\penalty0 (4):\penalty0 603--616, 2020.
\newblock ISSN 08966273.
\newblock \doi{10.1016/j.neuron.2020.06.014}.
\newblock URL
  \url{https://linkinghub.elsevier.com/retrieve/pii/S0896627320304682}.

\bibitem[Calvano et~al.(2020)Calvano, Calzolari, Denicol{\`o}, Harrington, and
  Pastorello]{Calvano2020}
Emilio Calvano, Giacomo Calzolari, Vincenzo Denicol{\`o}, Joseph~E Harrington,
  and Sergio Pastorello.
\newblock Protecting consumers from collusive prices due to {AI}.
\newblock \emph{Science}, 370\penalty0 (6520):\penalty0 1040--1042, 2020.
\newblock \doi{10.1126/science.abe3796}.
\newblock URL \url{https://www.science.org/doi/10.1126/science.abe3796}.

\bibitem[Sutton and Barto(2018)]{sutton2018reinforcement}
Richard~S Sutton and Andrew~G Barto.
\newblock \emph{Reinforcement learning: An introduction}.
\newblock MIT press, 2018.

\bibitem[Rummery and Niranjan(1994)]{Rummery1994}
Gavin~A Rummery and Mahesan Niranjan.
\newblock \emph{On-line Q-learning using connectionist systems}, volume~37.
\newblock Citeseer, 1994.

\bibitem[Sutton(1995)]{Sutton1995}
Richard~S Sutton.
\newblock Generalization in reinforcement learning: Successful examples using
  sparse coarse coding.
\newblock \emph{Advances in neural information processing systems}, 8, 1995.

\bibitem[Press and Dyson(2012)]{PressDyson2012}
William~H. Press and Freeman~J. Dyson.
\newblock Iterated {{Prisoner}}'s {{Dilemma}} contains strategies that dominate
  any evolutionary opponent.
\newblock \emph{Proceedings of the National Academy of Sciences}, 109\penalty0
  (26):\penalty0 10409--10413, 2012.
\newblock ISSN 0027-8424, 1091-6490.
\newblock \doi{10.1073/pnas.1206569109}.
\newblock URL \url{https://pnas.org/doi/full/10.1073/pnas.1206569109}.

\bibitem[Usui and Ueda(2021)]{UsuiUeda2021}
Yuki Usui and Masahiko Ueda.
\newblock Symmetric equilibrium of multi-agent reinforcement learning in
  repeated prisoner's dilemma.
\newblock \emph{Applied Mathematics and Computation}, 409:\penalty0 126370,
  2021.
\newblock ISSN 0096-3003.
\newblock \doi{10.1016/j.amc.2021.126370}.
\newblock URL
  \url{https://www.sciencedirect.com/science/article/pii/S0096300321004598}.

\bibitem[Meylahn et~al.(2022)Meylahn, Janssen, et~al.]{Meylahn2021}
Janusz~M Meylahn, Lars Janssen, et~al.
\newblock Limiting dynamics for {Q}-learning with memory one in symmetric
  two-player, two-action games.
\newblock \emph{Complexity}, 2022, 2022.
\newblock \doi{10.1155/2022/4830491}.
\newblock URL \url{https://doi.org/10.1155/2022/4830491}.

\bibitem[Barfuss et~al.(2019)Barfuss, Donges, and Kurths]{Barfuss2019}
W.~Barfuss, J.F. Donges, and J.~Kurths.
\newblock Deterministic limit of temporal difference reinforcement learning for
  stochastic games.
\newblock \emph{Physical Review E}, 99\penalty0 (043305), 2019.
\newblock URL \url{https://doi.org/10.1103/PhysRevE.99.043305}.

\bibitem[Barfuss(2020)]{Barfuss2020}
Wolfram Barfuss.
\newblock Reinforcement learning dynamics in the infinite memory limit.
\newblock In \emph{Proceedings of the 19th International Conference on
  Autonomous Agents and MultiAgent Systems}, pages 1768--1770, 2020.

\bibitem[Barfuss(2022)]{Barfuss2022}
Wolfram Barfuss.
\newblock Dynamical systems as a level of cognitive analysis of multi-agent
  learning.
\newblock \emph{Neural Computing and Applications}, 34\penalty0 (3):\penalty0
  1653--1671, 2022.
\newblock URL \url{https://doi.org/10.1007/s00521-021-06117-0}.

\bibitem[Lange et~al.(2012)Lange, Gabel, and Riedmiller]{lange2012batch}
Sascha Lange, Thomas Gabel, and Martin Riedmiller.
\newblock Batch reinforcement learning.
\newblock In \emph{Reinforcement learning}, pages 45--73. Springer, 2012.
\newblock URL \url{https://doi.org/10.1007/978-3-642-27645-3_2}.

\bibitem[Lin(1992)]{Lin1992}
Long-Ji Lin.
\newblock Self-improving reactive agents based on reinforcement learning,
  planning and teaching.
\newblock \emph{Machine Learning}, 8\penalty0 (3):\penalty0 293--321, 1992.
\newblock ISSN 1573-0565.
\newblock \doi{10.1007/BF00992699}.
\newblock URL \url{https://doi.org/10.1007/BF00992699}.

\bibitem[Mnih et~al.(2015)Mnih, Kavukcuoglu, Silver, Rusu, Veness, Bellemare,
  Graves, Riedmiller, Fidjeland, Ostrovski, Petersen, Beattie, Sadik,
  Antonoglou, King, Kumaran, Wierstra, Legg, and Hassabis]{MnihEtAl2015}
Volodymyr Mnih, Koray Kavukcuoglu, David Silver, Andrei~A. Rusu, Joel Veness,
  Marc~G. Bellemare, Alex Graves, Martin Riedmiller, Andreas~K. Fidjeland,
  Georg Ostrovski, Stig Petersen, Charles Beattie, Amir Sadik, Ioannis
  Antonoglou, Helen King, Dharshan Kumaran, Daan Wierstra, Shane Legg, and
  Demis Hassabis.
\newblock Human-level control through deep reinforcement learning.
\newblock \emph{Nature}, 518\penalty0 (7540):\penalty0 529--533, 2015.
\newblock ISSN 1476-4687.
\newblock \doi{10.1038/nature14236}.
\newblock URL \url{https://www.nature.com/articles/nature14236}.

\bibitem[Van~Seijen and Sutton(2015)]{VanSeijenSutton2015}
Harm Van~Seijen and Richard~S. Sutton.
\newblock A deeper look at planning as learning from replay.
\newblock In \emph{Proceedings of the 32nd International Conference on
  International Conference on Machine Learning - Volume 37}, ICML'15, page
  2314–2322. JMLR.org, 2015.

\bibitem[Wilson(1927)]{wilson1927probable}
Edwin~B Wilson.
\newblock Probable inference, the law of succession, and statistical inference.
\newblock \emph{Journal of the American Statistical Association}, 22\penalty0
  (158):\penalty0 209--212, 1927.
\newblock URL
  \url{https://www.tandfonline.com/doi/abs/10.1080/01621459.1927.10502953}.

\bibitem[Bialek(2012)]{Bialek2012}
William~S. Bialek.
\newblock \emph{Biophysics: searching for principles}.
\newblock Princeton University Press, Princeton, NJ, 2012.
\newblock ISBN 978-0-691-13891-6.

\bibitem[Vukov et~al.(2006)Vukov, Szab{\'o}, and Szolnoki]{VukovEtAl2006}
Jeromos Vukov, Gy{\"o}rgy Szab{\'o}, and Attila Szolnoki.
\newblock Cooperation in the noisy case: {{Prisoner}}'s dilemma game on two
  types of regular random graphs.
\newblock \emph{Physical Review E}, 73\penalty0 (6):\penalty0 067103, 2006.
\newblock \doi{10.1103/PhysRevE.73.067103}.
\newblock URL \url{https://link.aps.org/doi/10.1103/PhysRevE.73.067103}.

\bibitem[Szolnoki et~al.(2009)Szolnoki, Vukov, and Szab{\'o}]{SzolnokiEtAl2009}
Attila Szolnoki, Jeromos Vukov, and Gy{\"o}rgy Szab{\'o}.
\newblock Selection of noise level in strategy adoption for spatial social
  dilemmas.
\newblock \emph{Physical Review E}, 80\penalty0 (5):\penalty0 056112, 2009.
\newblock \doi{10.1103/PhysRevE.80.056112}.
\newblock URL \url{https://link.aps.org/doi/10.1103/PhysRevE.80.056112}.

\bibitem[Leonardos and Piliouras(2021)]{LeonardosPiliouras2021}
Stefanos Leonardos and Georgios Piliouras.
\newblock Exploration-{{Exploitation}} in {{Multi-Agent Learning}}:
  {{Catastrophe Theory Meets Game Theory}}.
\newblock \emph{Proceedings of the AAAI Conference on Artificial Intelligence},
  35\penalty0 (13):\penalty0 11263--11271, 2021.
\newblock ISSN 2374-3468.
\newblock \doi{10.1609/aaai.v35i13.17343}.
\newblock URL \url{https://ojs.aaai.org/index.php/AAAI/article/view/17343}.

\bibitem[Jia et~al.(2010)Jia, Liu, Yang, and Wang]{JiaEtAl2010}
Chun-Xiao Jia, Run-Ran Liu, Han-Xin Yang, and Bing-Hong Wang.
\newblock Effects of fluctuations on the evolution of cooperation in the
  prisoner's dilemma game.
\newblock \emph{Europhysics Letters}, 90\penalty0 (3):\penalty0 30001, May
  2010.
\newblock ISSN 0295-5075.
\newblock \doi{10.1209/0295-5075/90/30001}.
\newblock URL
  \url{https://iopscience.iop.org/article/10.1209/0295-5075/90/30001/meta}.

\bibitem[Assaf et~al.(2013)Assaf, Mobilia, and Roberts]{AssafEtAl2013}
Michael Assaf, Mauro Mobilia, and Elijah Roberts.
\newblock Cooperation {Dilemma} in {Finite} {Populations} under {Fluctuating}
  {Environments}.
\newblock \emph{Physical Review Letters}, 111\penalty0 (23):\penalty0 238101,
  December 2013.
\newblock \doi{10.1103/PhysRevLett.111.238101}.
\newblock URL \url{https://link.aps.org/doi/10.1103/PhysRevLett.111.238101}.

\bibitem[Wang et~al.(2022)Wang, Jia, Zhang, Zhu, Perc, Shi, and
  Wang]{WangEtAl2022}
Lu~Wang, Danyang Jia, Long Zhang, Peican Zhu, Matja{\v z} Perc, Lei Shi, and
  Zhen Wang.
\newblock L\'evy noise promotes cooperation in the prisoner's dilemma game with
  reinforcement learning.
\newblock \emph{Nonlinear Dynamics}, 2022.
\newblock ISSN 1573-269X.
\newblock \doi{10.1007/s11071-022-07289-7}.
\newblock URL \url{https://doi.org/10.1007/s11071-022-07289-7}.

\bibitem[Santos et~al.(2021)Santos, Levin, and Vasconcelos]{SantosEtAl2021}
Fernando~P. Santos, Simon~A. Levin, and V{\'i}tor~V. Vasconcelos.
\newblock Biased perceptions explain collective action deadlocks and suggest
  new mechanisms to prompt cooperation.
\newblock \emph{iScience}, 24\penalty0 (4):\penalty0 102375, 2021.
\newblock ISSN 2589-0042.
\newblock \doi{10.1016/j.isci.2021.102375}.
\newblock URL
  \url{https://www.sciencedirect.com/science/article/pii/S2589004221003436}.

\bibitem[Barfuss and Mann(2022)]{BarfussMann2022}
Wolfram Barfuss and Richard~P. Mann.
\newblock Modeling the effects of environmental and perceptual uncertainty
  using deterministic reinforcement learning dynamics with partial
  observability.
\newblock \emph{Physical Review E}, 105\penalty0 (3):\penalty0 034409, 2022.
\newblock \doi{10.1103/PhysRevE.105.034409}.
\newblock URL \url{https://link.aps.org/doi/10.1103/PhysRevE.105.034409}.

\bibitem[Nowak et~al.(2004)Nowak, Sasaki, Taylor, and Fudenberg]{NowakEtAl2004}
Martin~A. Nowak, Akira Sasaki, Christine Taylor, and Drew Fudenberg.
\newblock Emergence of cooperation and evolutionary stability in finite
  populations.
\newblock \emph{Nature}, 428\penalty0 (6983), April 2004.
\newblock ISSN 1476-4687.
\newblock \doi{10.1038/nature02414}.
\newblock URL \url{https://www.nature.com/articles/nature02414}.

\bibitem[Galla(2009)]{Galla2009}
Tobias Galla.
\newblock Intrinsic {{Noise}} in {{Game Dynamical Learning}}.
\newblock \emph{Physical Review Letters}, 103\penalty0 (19):\penalty0 198702,
  2009.
\newblock ISSN 0031-9007, 1079-7114.
\newblock \doi{10.1103/PhysRevLett.103.198702}.
\newblock URL \url{https://link.aps.org/doi/10.1103/PhysRevLett.103.198702}.

\bibitem[Galla(2011)]{Galla2011}
Tobias Galla.
\newblock Cycles of cooperation and defection in imperfect learning.
\newblock \emph{Journal of Statistical Mechanics: Theory and Experiment},
  2011\penalty0 (08):\penalty0 N08001, 2011.
\newblock ISSN 1742-5468.
\newblock \doi{10.1088/1742-5468/2011/08/N08001}.
\newblock URL \url{https://doi.org/10.1088/1742-5468/2011/08/n08001}.

\bibitem[Bladon and Galla(2011)]{BladonGalla2011}
Alex~J. Bladon and Tobias Galla.
\newblock Learning dynamics in public goods games.
\newblock \emph{Physical Review E}, 84\penalty0 (4):\penalty0 041132, 2011.
\newblock \doi{10.1103/PhysRevE.84.041132}.
\newblock URL \url{https://link.aps.org/doi/10.1103/PhysRevE.84.041132}.

\bibitem[Panait et~al.(2008)Panait, Tuyls, and Luke]{PanaitEtAl2008}
Liviu Panait, Karl Tuyls, and Sean Luke.
\newblock Theoretical advantages of lenient learners: {{An}} evolutionary game
  theoretic perspective.
\newblock \emph{The Journal of Machine Learning Research}, 9:\penalty0
  423--457, 2008.

\bibitem[Barfuss et~al.(2017)Barfuss, Donges, Wiedermann, and
  Lucht]{BarfussEtAl2017}
Wolfram Barfuss, Jonathan~F. Donges, Marc Wiedermann, and Wolfgang Lucht.
\newblock Sustainable use of renewable resources in a stylized
  social\textendash ecological network model under heterogeneous resource
  distribution.
\newblock \emph{Earth System Dynamics}, 8\penalty0 (2):\penalty0 255--264,
  2017.
\newblock URL \url{https://doi.org/10.5194/esd-8-255-2017}.

\bibitem[Geier et~al.(2019)Geier, Barfuss, Wiedermann, Kurths, and
  Donges]{GeierEtAl2019}
Fabian Geier, Wolfram Barfuss, Marc Wiedermann, J{\"u}rgen Kurths, and
  Jonathan~F. Donges.
\newblock The physics of governance networks: Critical transitions in contagion
  dynamics on multilayer adaptive networks with application to the sustainable
  use of renewable resources.
\newblock \emph{The European Physical Journal Special Topics}, 228\penalty0
  (11):\penalty0 2357--2369, 2019.
\newblock ISSN 1951-6401.
\newblock \doi{10.1140/epjst/e2019-900120-4}.
\newblock URL \url{https://doi.org/10.1140/epjst/e2019-900120-4}.

\bibitem[Barfuss et~al.(2018)Barfuss, Donges, Lade, and
  Kurths]{BarfussEtAl2018}
Wolfram Barfuss, Jonathan~F. Donges, Steven~J. Lade, and J{\"u}rgen Kurths.
\newblock When optimization for governing human-environment tipping elements is
  neither sustainable nor safe.
\newblock \emph{Nature Communications}, 9\penalty0 (1):\penalty0 2354, 2018.
\newblock ISSN 2041-1723.
\newblock \doi{10.1038/s41467-018-04738-z}.
\newblock URL \url{https://www.nature.com/articles/s41467-018-04738-z}.

\bibitem[Barfuss et~al.(2020)Barfuss, Donges, Vasconcelos, Kurths, and
  Levin]{BarfussEtAl2020}
Wolfram Barfuss, Jonathan~F. Donges, V{\'i}tor~V. Vasconcelos, J{\"u}rgen
  Kurths, and Simon~A. Levin.
\newblock Caring for the future can turn tragedy into comedy for long-term
  collective action under risk of collapse.
\newblock \emph{Proceedings of the National Academy of Sciences}, 117\penalty0
  (23):\penalty0 12915--12922, 2020.
\newblock \doi{10.1073/pnas.1916545117}.
\newblock URL \url{https://www.pnas.org/doi/abs/10.1073/pnas.1916545117}.

\bibitem[den Boer et~al.(2022)den Boer, Meylahn, and Schinkel]{denBoer2022}
Arnoud~V. den Boer, Janusz~M. Meylahn, and Maarten~Pieter Schinkel.
\newblock Artificial collusion: Examining supra-competitive pricing by
  autonomous {Q}-learning algorithms.
\newblock Available at SSRN, 2022.
\newblock URL \url{https://ssrn.com/abstract=4213600}.

\bibitem[Wunder et~al.(2010)Wunder, Littman, and Babes]{WunderEtAl2010}
Michael Wunder, Michael~L. Littman, and Monica Babes.
\newblock Classes of multiagent q-learning dynamics with epsilon-greedy
  exploration.
\newblock In \emph{ICML}, pages 1167--1174, 2010.
\newblock URL \url{https://icml.cc/Conferences/2010/papers/191.pdf}.

\end{thebibliography}


\newpage
\appendix

\renewcommand{\thesection}{Supplementary Information \arabic{section}}    

\section*{Supplementary Information}
\renewcommand{\thefigure}{SI~\arabic{figure}}
\setcounter{figure}{0}
\renewcommand{\thetable}{SI~\arabic{table}}
\setcounter{table}{0}
\label{app:SI}
\added{Here we provide the supplementary information for the manuscript: ``Intrinsic fluctuations of reinforcement learning promote cooperation''. In Figure \ref{fig:Robustness_Environment_T15_S02}, we plot the results of our robustness analysis for the environmental parameters $T=1.5$ and $S=-0.2$, and in Figure \ref{fig:Robustness_Environment_T125_S025} we plot the results of our robustness analysis for the environmental parameters $T=1.25$ and $S=-0.25$. We see that levels of cooperation close to one are possible for both environments. These can be achieved on relatively short time scales compared to the timescale required by the algorithm without batches. }

\begin{sidewaysfigure}[ht]
    \includegraphics[width=\textwidth]{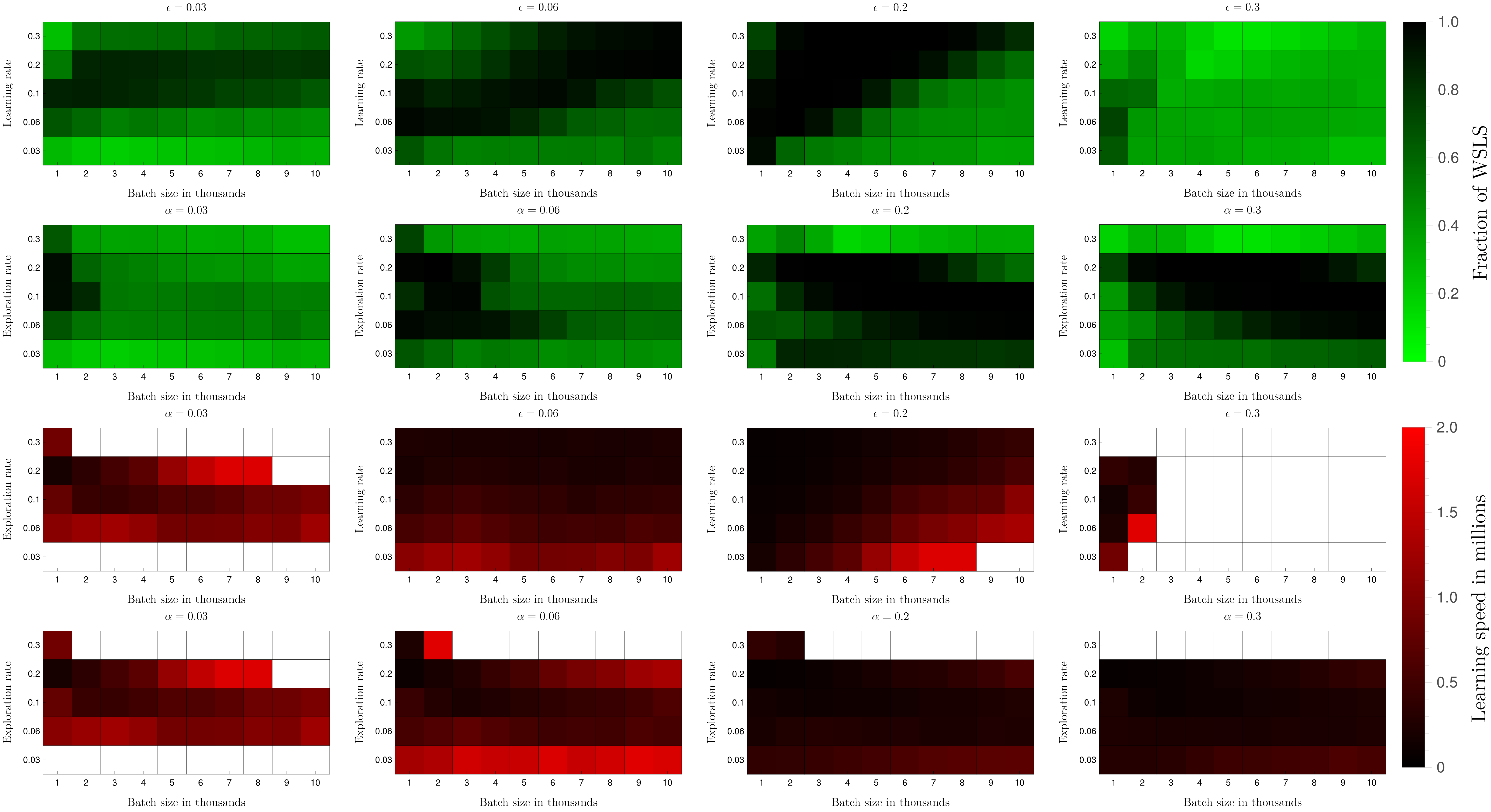}
    \caption{\added{Plots showing the robustness of the results for different parameter values. The top two rows show the fraction of trajectories (1000 samples) that end in the WSLS strategy pair at time $2\times 10^{6}$. The bottom two rows show the time it takes for the fraction of trajectories in the WSLS strategy pair to reach $0.4$ in millions of time steps (we use white to represent trajectories that never reached $0.4$). The x-axis always represents the batch size in thousands, and the y-axis represents either the learning rate $\alpha$ or the exploration rate $\epsilon$. In all cases, we set $\delta=0.99$.}}
    \label{fig:Robustness_Environment_T15_S02}
\end{sidewaysfigure}

\begin{sidewaysfigure}[ht]
    \includegraphics[width=\textwidth]{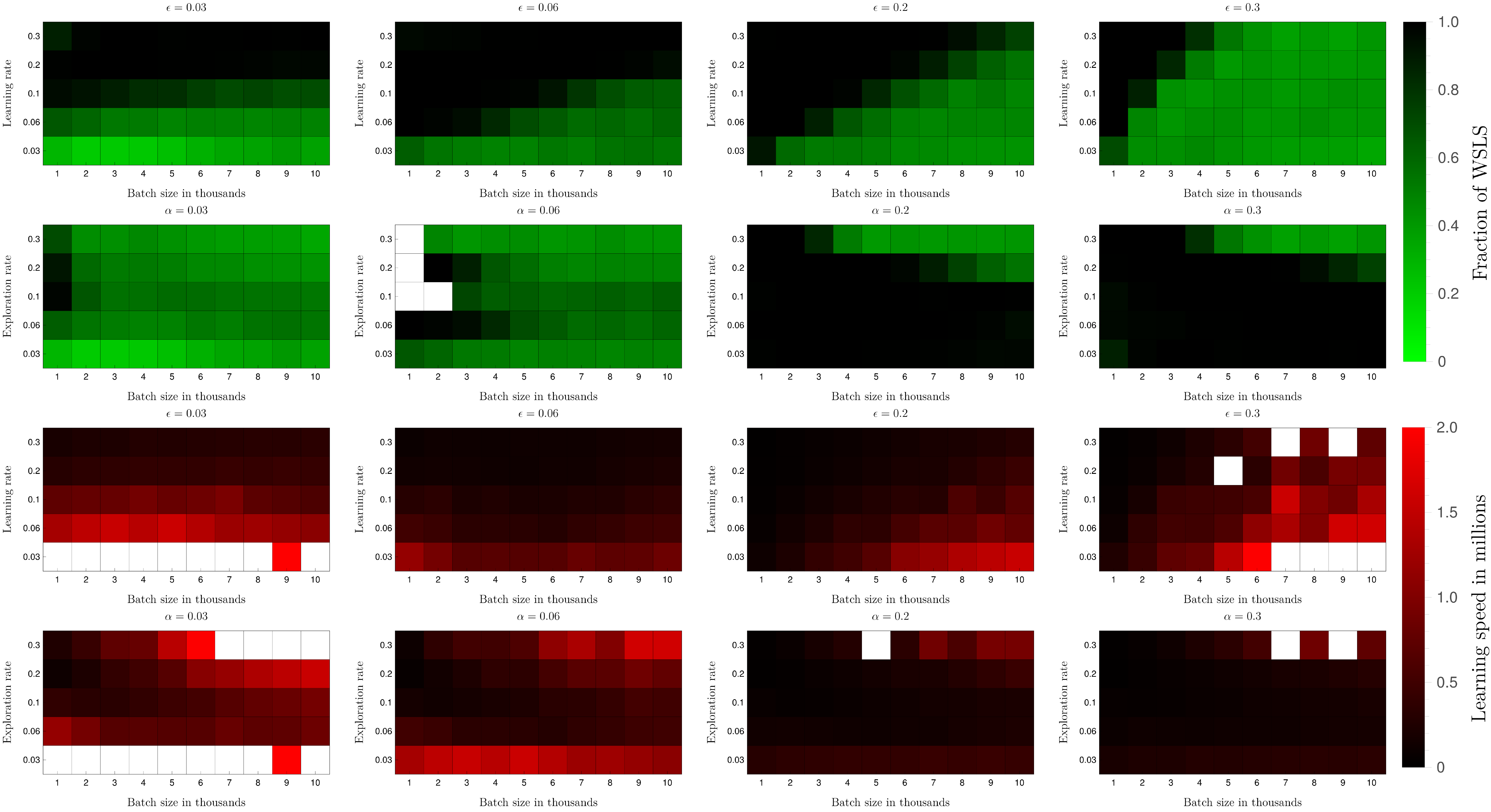}
    \caption{\added{Plots showing the robustness of the results for different parameter values. The top two rows show the fraction of trajectories (1000 samples) that end in the WSLS strategy pair at time $2\times 10^{6}$. The bottom two rows show the time it takes for the fraction of trajectories in the WSLS strategy pair to reach $0.4$ in millions of time steps (we use white to represent trajectories that never reached $0.4$). The x-axis always represents the batch size in thousands, and the y-axis represents either the learning rate $\alpha$ or the exploration rate $\epsilon$. All plots are for the environment with $T=1.25$ and $S=-0.25$. In all cases, we set $\delta=0.99$.}}
    \label{fig:Robustness_Environment_T125_S025}
\end{sidewaysfigure}

\end{document}